%% file: main.tex
\documentclass[10pt,twocolumn,letterpaper]{article}

\usepackage[OT1]{fontenc}
\usepackage[pagenumbers]{wacv}

\input{preamble}

\newcommand{\paperTitle}{SHIN: Scalable Hierarchy for INcremental 3D Gaussian Splatting}
\definecolor{linkblue}{rgb}{0.21,0.49,0.74}
\usepackage[breaklinks,colorlinks,allcolors=linkblue]{hyperref}
\hypersetup{
  pdfauthor={Reagan Koo, Yeong-Gyu Kim, Eun-Seok Ryu},
  pdftitle={\paperTitle},
  pdfsubject={Scalable 3D Gaussian Splatting},
  pdfkeywords={3D Gaussian Splatting, scalable representation, progressive transmission}
}

\title{\paperTitle}

\author{Reagan Koo \qquad Yeong-Gyu Kim \qquad Eun-Seok Ryu\\
Sungkyunkwan University\\
{\tt\small \{reagan9, kyk2798, esryu\}@skku.edu}
}

\begin{document}
\maketitle
\input{sec/0_abstract}
\input{sec/1_intro}
\newpage
\input{sec/2_formatting}
\newpage
\input{sec/3_method}
\input{sec/3_experimental_results}
\input{sec/3_ablation}
\FloatBarrier
\input{sec/4_conclusion}
\vfill
{
    \small
    \bibliographystyle{ieeenat_fullname}
    \bibliography{main}
}

\clearpage
\appendix
\raggedbottom
\twocolumn[
\begin{center}
{\LARGE\bfseries \paperTitle\par}
\vspace{0.5em}
{\large\bfseries Supplementary Material\par}
\vspace{1.0em}
\end{center}
]
\input{sec/supplementary_implementation}

\end{document}

%% file: preamble.tex
\usepackage{booktabs}
\usepackage{multirow}
\usepackage{makecell}
\usepackage{adjustbox}
\usepackage{colortbl}
\usepackage{placeins}
\usepackage{dblfloatfix}
\usepackage{algorithm}
\usepackage{algpseudocode}

%% file: sec/0_abstract.tex
\begin{abstract}
3D Gaussian Splatting (3DGS) provides high-quality real-time novel-view synthesis, but deployment budgets vary across devices and network conditions. Most compact methods target one operating point, so another budget requires a separate representation. SHIN is a scalable 3DGS hierarchy with a reusable Base and additive enhancement layers, so each higher-quality operating point extends rather than replaces the decoded representation. Constructing this hierarchy requires deciding which Gaussians each stage adds and how much spherical-harmonic (SH) appearance each retains, because these choices remain fixed in later prefixes. SHIN derives one shared Gaussian order from multi-scale position--scale sensitivity and divides each stage budget between adding Gaussians and retaining SH bands using calibrated gradient-based scores per byte. Lightweight opacity updates adjust compositing while transmitted geometry and SH coefficients remain unchanged. Across three benchmarks, SHIN progressively improves one cumulative representation and achieves the best PSNR--size efficiency in every overlapping comparison with the multi-rate and progressive baselines.
\end{abstract}

%% file: sec/1_intro.tex
\section{Introduction}
\label{sec:intro}

\begingroup
\setlength{\parskip}{0pt}

3D Gaussian Splatting (3DGS) provides high-quality real-time novel-view synthesis \cite{kerbl3Dgaussians}, but its large memory footprint has motivated compact representations optimized for a prescribed target size \cite{fan2024lightgaussian,hanson2025pup,niedermayr2024compressed,Girish_EAGLES_ECCV2024,hanson2025speedy}. These methods are effective when the deployment budget is known in advance. In practice, however, the same scene may serve devices with different storage capacities, and the available network bandwidth may change over time. Maintaining a separate asset for every target budget duplicates storage, while moving an existing client to a higher-quality point requires transferring a replacement representation.

This paper presents SHIN, a scalable 3DGS hierarchy built from one reusable asset (\cref{fig:scalable_overview}(a)). A Base is transmitted once, and each higher operating point is reached by appending an enhancement layer, so previously decoded spatial and appearance records remain useful as the available budget increases \cite{shi2025lapisgs}. This cumulative requirement differs from Level-of-Detail (LoD; \cref{fig:scalable_overview}(b)), which may expose multiple rendering states from one shared representation without requiring those states to form an additive transmission sequence \cite{cheng2026clodgs,octree,seo2025flod,Shen_2025_CVPR}. The distinction is not the number of operating points, but whether moving between them preserves the data already delivered.

\begin{figure}[!t]
    \centering
    \includegraphics[width=\columnwidth]{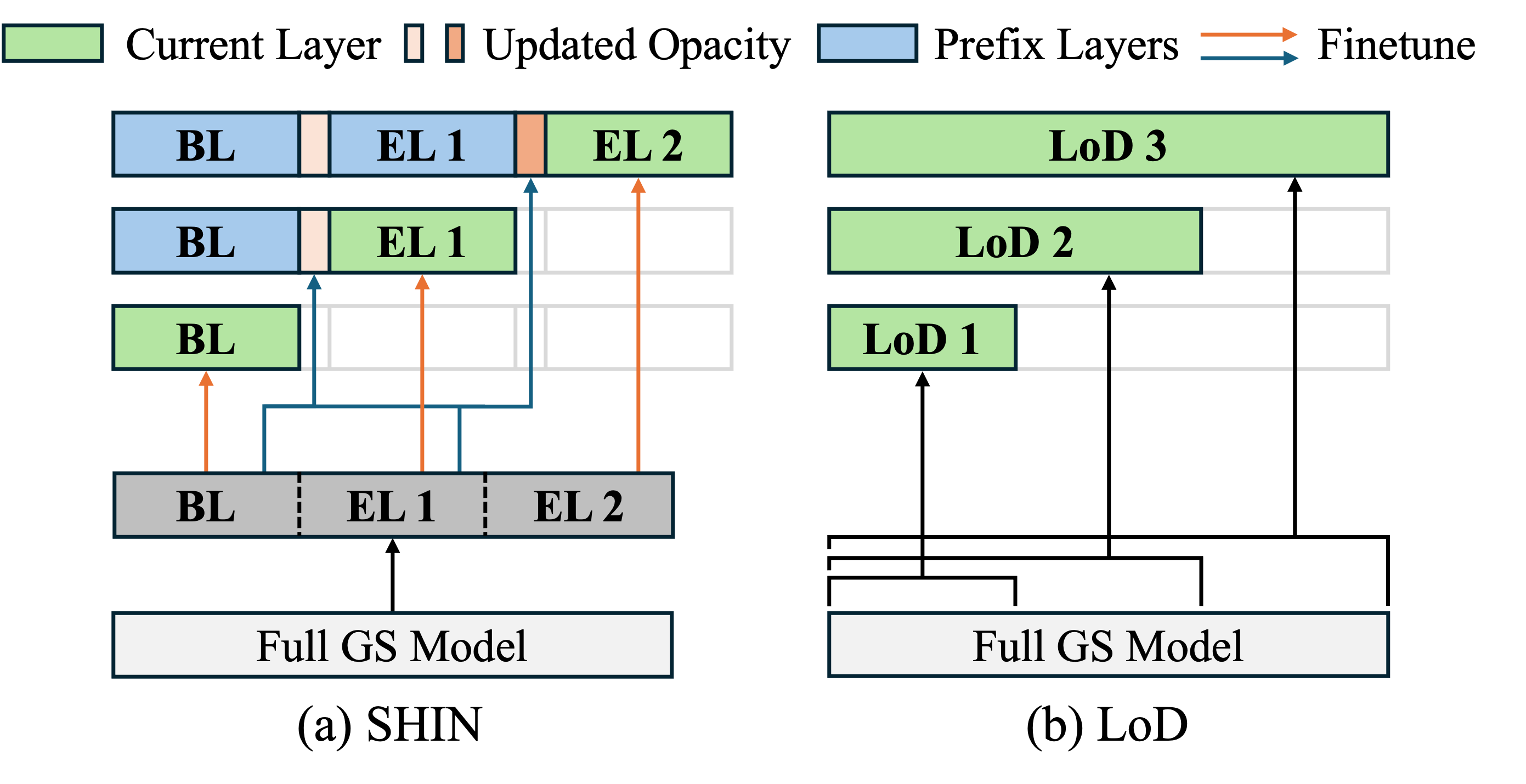}
    \caption{\textbf{Scalable hierarchy versus LoD.} \textbf{(a)} SHIN cumulatively extends a decoded Base with enhancement layers. \textbf{(b)} LoD selects operating states from a shared representation without requiring cumulative transmission.}
    \label{fig:scalable_overview}
\end{figure}

Cumulative reuse requires the Gaussian contents of successive operating points to be nested. If each budget independently selects its own subset, a Gaussian used by the Base may disappear at a higher rate, forcing previously decoded data to be replaced. SHIN enforces this nesting with one shared Gaussian order, so that every operating point extends the same prefix rather than redefining it. Ordered-prefix representations provide this nesting along the Gaussian-count axis \cite{guo2026matryoshka}.

Gaussian count is only one dimension of 3DGS storage. Each Gaussian also carries view-dependent spherical-harmonic (SH) coefficients, and retaining additional bands increases the size of its record. Under a limited stage budget, richer SH appearance for existing Gaussians leaves fewer bytes for additional spatial support, whereas lower retained degrees allow more Gaussians to be added but restrict their appearance.

A shared order guarantees nested Gaussian contents, but it does not determine how much SH representation is stored with each Gaussian. The hierarchy must therefore decide both which Gaussians are added at each stage and how much SH representation each retains. These choices affect all later operating points: spending early bytes on higher-order SH can delay additional spatial support, whereas retaining too little SH can limit later view-dependent refinement. Because transmitted records are reused by all subsequent prefixes, the stages cannot be optimized independently. This coupled decision is a persistent allocation problem.

Neither the Base nor the largest operating point can be designed in isolation. Optimizing only the Base may spend early bytes on choices that later stages cannot revise, while optimizing the largest point first can leave shorter prefixes poorly balanced. A useful cumulative hierarchy must therefore maintain strong reconstruction throughout the sequence, not only at its final size. This makes the interaction between stage budgets and persistent Gaussian records central to the construction, rather than a post-hoc truncation of one endpoint model.

The shared order must remain useful across the full prefix range: early prefixes should establish broad scene structure, whereas later prefixes should add Gaussians that resolve finer residuals. This motivates measuring structural sensitivity across multiple image scales before constructing the order.

SHIN converts a 3DGS model into a reusable Base and additive enhancement layers. One shared Gaussian order keeps the stage contents nested, while Layer--Degree Allocation divides each stage budget between adding the next Gaussian and retaining additional SH bands. Since newly added Gaussians also change transmittance, lightweight opacity updates adjust compositing while previously transmitted geometry and SH coefficients remain unchanged.

The contributions are summarized as follows:
\begin{enumerate}
    \item A formulation of cumulative 3DGS construction as a persistent allocation problem in which adding Gaussians and selecting their retained SH degrees jointly determine a sequence of reusable operating points.
    \item Layer--Degree Allocation, which combines a shared multi-scale Gaussian order with calibrated gradient-based scores per byte to allocate each stage budget between spatial support and retained SH appearance.
    \item A cumulative refinement and decoding scheme that improves reconstruction with additive enhancement layers while reusing transmitted Gaussian records.
\end{enumerate}

\endgroup

%% file: sec/2_formatting.tex
\section{Related Work}

\noindent\textbf{Compact 3D Gaussian representations.}
Compact methods reduce primitive count or attribute storage. LightGaussian and PUP estimate pruning importance \cite{fan2024lightgaussian,hanson2025pup}; Compact3DGS, LP-3DGS, and MaskGaussian learn Gaussian masks or existence \cite{lee2024c3dgs,zhang2024lp,liu2025maskgaussian}; and GaussianSpa and GS$^2$ optimize sparsification or spatial redistribution \cite{zhang2025gaussianspa,yang2026gs2}. SteepGS revisits density control for compact representations \cite{Wang_2025_CVPR}. Beyond primitive reduction, Compressed3DGS and EAGLES employ compact attribute encodings \cite{niedermayr2024compressed,Girish_EAGLES_ECCV2024}, while HAC and ContextGS exploit spatial or hierarchical context for entropy modeling \cite{Chen_HAC_ECCV2024,Wang_ContextGS_NeurIPS2024}. RDO-Gaussian jointly optimizes Gaussian pruning, adaptive SH pruning, and quantization under a rate--distortion objective \cite{wang2024rdogaussian}, while FCGS enables feed-forward Gaussian compression \cite{Chen_FCGS_ICLR2025}. These methods target individual operating points. SHIN instead considers cumulative construction, where choices about adding Gaussians and retaining appearance persist in later prefixes.

\noindent\textbf{Flexible and continuous 3DGS representations.}
Beyond a single compact operating point, flexible methods expose multiple model sizes or rendering budgets from one learned representation. FlexGS adapts Gaussian subsets and attributes to a requested model size \cite{Liu_2025_CVPR}, Learning Differentiable Hierarchies learns per-Gaussian hierarchy values for variable-size active subsets \cite{Pan2026learning}, and CLoD-GS continuously modulates Gaussian contributions according to viewing conditions \cite{cheng2026clodgs}. Matryoshka Gaussian Splatting (MGS) learns an ordered Gaussian representation in which any primitive prefix yields a coherent reconstruction, enabling continuous LoD by varying the prefix length \cite{guo2026matryoshka}. These formulations do not address the joint decision of when a Gaussian enters the representation and how much appearance capacity it retains under cumulative reuse.

\begin{figure*}[!t]
\centering
\includegraphics[width=\textwidth]{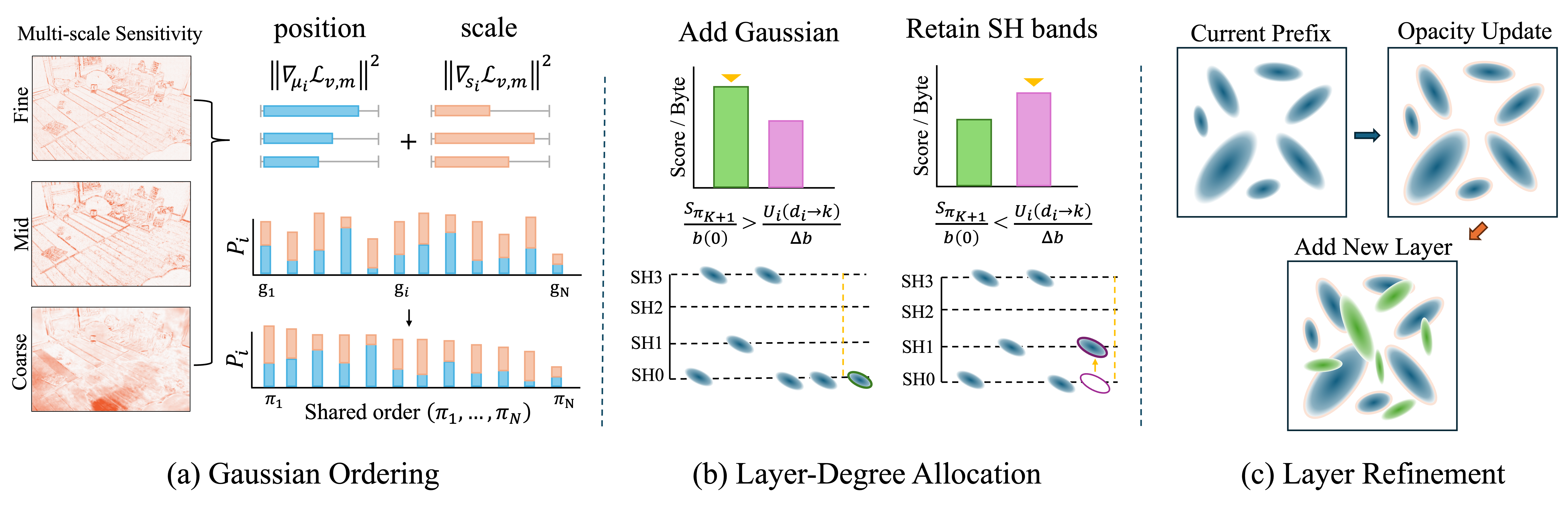}
\caption{\textbf{SHIN pipeline.} \textbf{(a)} Multi-scale position--scale sensitivity defines a Gaussian order shared across stages. \textbf{(b)} Layer--Degree Allocation divides each stage budget between adding Gaussians and retaining SH bands using calibrated scores per byte. \textbf{(c)} Each layer is appended while transmitted geometry and SH remain fixed, with opacity updates applied to the existing prefix.}
\label{fig:method_overview}
\end{figure*}

\noindent\textbf{Hierarchical, LoD, and progressive 3DGS representations.}
Octree-GS extends Scaffold-GS anchors~\cite{scaffoldgs} into a multi-resolution hierarchy and selects LoD anchors according to viewing condition~\cite{octree}. FLoD, LOD-GS, and LODGE provide scale- or spatially structured LoDs \cite{seo2025flod,Shen_2025_CVPR,kulhanek2025lodge}, while PRoGS prioritizes splats for progressive partial loading \cite{Zoomers_2025_WACV}. GoDe assigns Gaussians to progressive levels through iterative gradient-informed masking under a prescribed Gaussian-count schedule \cite{di2025gode}. L3GS constructs prescribed splat-count layers and schedules their object- and layer-level delivery from predicted viewport and bandwidth \cite{tsai2025l3gs}. LapisGS constructs Base and enhancement layers and allows preceding opacity values to adapt as new Gaussians are added \cite{shi2025lapisgs}. These methods structure progressive access or delivery, but do not allocate a common cumulative byte budget between Gaussian addition and retained per-Gaussian SH capacity.

%% file: sec/3_method.tex
\section{Method}
\label{sec:method}

\noindent\textbf{3DGS Background and Method Overview.}

A 3D Gaussian Splatting (3DGS) scene comprises anisotropic Gaussians $\mathcal G=\{g_i\}_{i=1}^{N}$. Each $g_i$ contains a center $\boldsymbol\mu_i\in\mathbb R^3$, scale $\mathbf s_i\in\mathbb R^3$, rotation $\mathbf q_i$, opacity logit $o_i$, and spherical-harmonic (SH) coefficients $\mathbf c_i$. Scale and rotation define its 3D covariance, while the SH coefficients determine view-dependent color. Higher SH degrees increase appearance capacity at additional storage cost.

Rendering projects the Gaussians to the image plane and alpha-composites them in depth order. At pixel $x$, the rendered color is
\begin{equation}
\mathbf C(x)=\sum_i T_i(x)\alpha_i(x)\mathbf c_i(x),
\qquad
T_i(x)=\prod_{j<i}\bigl(1-\alpha_j(x)\bigr),
\label{eq:alpha-compositing}
\end{equation}
where $\alpha_i(x)$ combines the projected Gaussian footprint with $\sigma(o_i)$. For cumulative construction, model size depends jointly on Gaussian count and SH degree, while adding Gaussians changes the transmittance of the existing prefix. A scalable hierarchy must therefore allocate geometry and appearance together and account for cross-layer compositing.

Cumulative reuse creates three requirements. First, every stage must preserve the Gaussian contents already transmitted, so all stages must extend one shared order. Second, storage depends on both Gaussian count and retained SH degree, so the stage budget must be divided between adding Gaussians and retaining appearance. Third, newly added Gaussians change transmittance, so compositing must be adjusted without changing earlier geometry or SH coefficients. \Cref{fig:method_overview} summarizes how SHIN addresses these requirements.

\subsection{Cumulative Hierarchy}
\label{sec:problem-formulation}

SHIN stores each Gaussian once and updates only its opacity thereafter. Each source Gaussian $i$ is therefore assigned a layer $\ell_i\in\{0,\ldots,T\}$ and a retained SH degree $d_i\in\{0,1,2,3\}$. The degree determines which appearance coefficients are stored with that Gaussian: all SH coefficients through degree $d_i$ are included, and later layers do not alter this record. The layer contents and cumulative prefix are
\begin{equation}
\mathcal H_t=\{i:\ell_i=t\},
\qquad
\mathcal P_t=\bigcup_{k=0}^{t}\mathcal H_k.
\label{eq:prefix-sets}
\end{equation}
Here, $\mathcal H_t$ contains the Gaussians added in layer $t$, and $\mathcal P_t$ is the cumulative prefix available at that operating point. The Base is $\mathcal P_0$, and enhancement $t$ transmits $\mathcal H_t$ together with an opacity update described in \cref{sec:method-refinement}. Once Gaussian $i$ is transmitted, its layer, geometry, and retained SH coefficients remain fixed; only its opacity may be updated. \Cref{fig:per_layer_effect} visualizes the distinction between the disjoint layer contents $\mathcal H_t$ and their cumulative prefixes $\mathcal P_t$.

The Gaussian payload of stage $t$ is
\begin{equation}
R_t^G=\sum_{i:\ell_i\leq t} b(d_i),
\label{eq:prefix-rate}
\end{equation}
where $b(d)$ denotes the storage cost of a Gaussian retained at degree $d$. The budgets $B_0^G<\cdots<B_T^G$ specify the desired cumulative Gaussian sizes. Separate opacity-update data accompany the enhancement layers and are included in the reported total model sizes, while $B_t^G$ controls the allocation of Gaussian data itself.

\paragraph{Cumulative constraints.}
SHIN imposes the following cumulative constraints:
\begin{equation}
R_t^G\leq B_t^G,
\qquad
\mathcal P_{t-1}\subseteq\mathcal P_t,
\qquad
\boldsymbol\vartheta_i^{(t)}=\boldsymbol\vartheta_i^{(\ell_i)}
\quad \forall\,t\geq\ell_i,
\label{eq:cumulative-constraints}
\end{equation}
where
\begin{equation}
\boldsymbol\vartheta_i=
\left(\boldsymbol\mu_i,\mathbf s_i,\mathbf q_i,
\mathbf c_i^{\leq d_i}\right)
\label{eq:persistent-record}
\end{equation}
collects the transmitted geometry and retained SH coefficients of Gaussian $i$. The first constraint enforces the cumulative payload budget, the second keeps successive operating points nested, and the third prevents previously transmitted geometry and SH records from changing. Opacity is excluded from $\boldsymbol\vartheta_i$ because it is adapted during enhancement refinement.

\input{sec/3_perlayer_float}

\subsection{Gaussian Ordering}
\label{sec:method-ordering}

The same Gaussian order is reused at every operating point. It must therefore place Gaussians that establish broad scene structure near the beginning while continuing to rank Gaussians that refine local detail later in the sequence. A score designed around one pruning cutoff does not explicitly account for both roles.

PUP derives structural importance from the position--scale Fisher block \cite{hanson2025pup}. SHIN retains this structural focus, but measures position--scale sensitivity over the coarse, mid, and fine bands of a Laplacian image pyramid before combining the evidence into one order. Each Gaussian is therefore evaluated by its influence on both broad structure and localized detail.

For training view $v$ and pyramid scale $m\in\{4,2,1\}$, let $\mathcal B_m$ denote the corresponding Laplacian-pyramid band and define
\begin{equation}
\mathcal L_{v,m}^{\mathrm{lap}}=
\left\|\mathcal B_m(\hat{\mathbf I}_v)
-\mathcal B_m(\mathbf I_v)\right\|_1.
\label{eq:laplacian-loss}
\end{equation}
The structural importance of Gaussian $i$ is
\begin{equation}
P_i=\sum_{v,m}
\left(
\left\|\frac{\partial\mathcal L_{v,m}^{\mathrm{lap}}}
{\partial\boldsymbol\mu_i}\right\|_2^2+
\left\|\frac{\partial\mathcal L_{v,m}^{\mathrm{lap}}}
{\partial\mathbf s_i}\right\|_2^2
\right),
\quad P_{\pi_1}\geq\cdots\geq P_{\pi_N}.
\label{eq:spatial-sensitivity}
\end{equation}
The pyramid does not assign frequency bands to individual layers. It exposes structural sensitivity at several image scales before the Gaussians are sorted. If the current prefix contains the first $K$ Gaussians, only $\pi_{K+1}$ can be added next. This guarantees that every operating point extends the same Gaussian prefix.

\begin{algorithm}[!t]
\caption{Layer--Degree Allocation}
\label{alg:layer_degree}
\scriptsize
\begin{algorithmic}[1]
\Require Gaussian order $\pi$, scores $S_i,U_i$, costs $b(\cdot)$, cumulative budgets $\{B_t^G\}_{t=0}^{T}$
\Ensure Layer sets $\{\mathcal H_t\}_{t=0}^{T}$ and retained degrees $\{d_i\}$
\State $K\gets0$; $R\gets0$
\For{$t=0,\ldots,T$}
    \State $\mathcal H_t\gets\emptyset$
    \While{at least one candidate fits $B_t^G-R$}
        \State Form feasible degree increases $\mathcal C_{\mathrm{degree}}=\{(i,d_i\!\rightarrow\!k):i\in\mathcal H_t,\ k>d_i,\ b(k)-b(d_i)\le B_t^G-R\}$
        \State $\mathcal C\gets\mathcal C_{\mathrm{degree}}$
        \If{$K<N$ and adding $\pi_{K+1}$ at degree 0 fits}
            \State Add this Gaussian candidate to $\mathcal C$
        \EndIf
        \State Select $c^\star\in\mathcal C$ with the highest score per byte
        \If{$c^\star$ adds a Gaussian}
            \State $K\gets K+1$
            \State $\mathcal H_t\gets\mathcal H_t\cup\{\pi_K\}$; $d_{\pi_K}\gets0$
            \State $R\gets R+b(0)$
        \Else
            \State Let $c^\star=(i,d_i\!\rightarrow\!k)$
            \State $R\gets R+b(k)-b(d_i)$; $d_i\gets k$
        \EndIf
    \EndWhile
\EndFor
\end{algorithmic}
\end{algorithm}

\input{sec/3_main_results_floats}

\subsection{Layer--Degree Allocation}
\label{sec:method-allocation}

Before a layer is transmitted, its remaining payload can be used in two ways: add the next Gaussian in the shared order at degree 0, or increase the retained SH degree of a Gaussian already assigned to that layer. A fixed Gaussian count cannot trade spatial coverage for richer appearance, while a fixed SH degree predetermines how many Gaussian records fit within a budget. SHIN therefore compares these two choices by score per byte.

The structural score is derived from the position--scale importance $P_i$. For higher-order appearance, an auxiliary scalar $h_{i,l}=1$ multiplies SH band $l\in\{1,2,3\}$ of Gaussian $i$. For a training view $v$, let
\begin{equation}
\mathcal E_v=\frac{1}{3HW}
\left\|\hat{\mathbf I}_v-\mathbf I_v\right\|_2^2.
\label{eq:reconstruction-error}
\end{equation}
Rendered and target RGB values are clamped to $[0,1]$ before evaluating $\mathcal E_v$.
The score for retaining bands from degree $d+1$ through $k$ is
\begin{equation}
U_i(d\!\rightarrow\!k)=
\sum_{l=d+1}^{k}\sum_v
\left(\frac{\partial\mathcal E_v}{\partial h_{i,l}}\right)^2.
\label{eq:degree-utility}
\end{equation}
Since SH0 is always retained, $U_i$ measures only higher-order appearance.

Because $P_i$ and $U_i$ are obtained from different perturbations, the structural score is rescaled as
\begin{equation}
S_i=\gamma P_i,
\qquad
\gamma=
\frac{Q_{0.95}\!\left(G^{\mathrm{op}}_{>0}\right)}
{\max\!\left(Q_{0.95}\!\left(P_{>0}\right),10^{-12}\right)}.
\label{eq:score-calibration}
\end{equation}
where $G^{\mathrm{op}}$ is the sensitivity to a temporary opacity gate and $>0$ denotes finite positive values. The calibrated scores provide a common score-per-byte ranking for the two feasible allocation actions.

\begin{equation}
\begin{aligned}
r_{\mathrm{add}}&=\frac{S_{\pi_{K+1}}}{b(0)},\\
r_{\mathrm{degree}}(i,k)&=
\frac{U_i(d_i\!\rightarrow\!k)}{b(k)-b(d_i)}.
\end{aligned}
\label{eq:action-score-per-byte}
\end{equation}
\Cref{alg:layer_degree} repeatedly selects the feasible candidate with the highest score per byte until no candidate fits the remaining cumulative budget. All retained degrees are finalized before the layer is transmitted, and later layers do not alter previously transmitted SH records.

\input{sec/3_main_results_table}
\input{sec/3_rate_transition_tables}
\input{sec/3_allocation_floats}

\subsection{Layer Refinement and Decoding}
\label{sec:method-refinement}

Each stage adapts the allocated records to the reconstruction objective without changing their payload. Let
\begin{equation}
\Theta_t^{\mathrm{new}}=
\left\{\boldsymbol\mu_i,\mathbf s_i,\mathbf q_i,o_i,
\mathbf c_i^{\leq d_i}:i\in\mathcal H_t\right\}
\label{eq:new-layer-parameters}
\end{equation}
denote the parameters of the Gaussians newly added at stage $t$. For an enhancement stage, SHIN additionally optimizes prefix-opacity deltas
\begin{equation}
\Delta O_t=
\left\{\Delta o_i^{(t)}:i\in\mathcal P_{t-1}\right\}.
\label{eq:prefix-opacity-deltas}
\end{equation}

The Base has no earlier decoded state to preserve. Its allocated records are refined by
\begin{equation}
\Theta_0^\star=
\arg\min_{\Theta_0^{\mathrm{new}}}
\frac{1}{|\mathcal V|}\sum_{v\in\mathcal V}
\mathcal L_{\mathrm{3DGS}}
\!\left(\hat{\mathbf I}_v(\mathcal P_0),\mathbf I_v\right),
\label{eq:base-refinement}
\end{equation}
without densification or degree changes, where $\mathcal L_{\mathrm{3DGS}}$ is the standard 3DGS photometric objective \cite{kerbl3Dgaussians}.

Enhancement stages cannot modify geometry or SH coefficients in the existing prefix without retransmission. SHIN therefore optimizes only the newly added Gaussian records and the opacity deltas of the existing prefix:
\begin{equation}
\begin{aligned}
\left(\Theta_t^{\mathrm{new}},\Delta O_t\right)^\star
={}&\arg\min_{\Theta_t^{\mathrm{new}},\Delta O_t}
\frac{1}{|\mathcal V|}\sum_{v\in\mathcal V}\\[-2pt]
&\left\|\hat{\mathbf I}_v(\mathcal P_t)-\mathbf I_v\right\|_1,
\qquad t\geq1.
\end{aligned}
\label{eq:enhancement-refinement}
\end{equation}
The loss is evaluated on the cumulative rendering $\mathcal P_t$, while the position, scale, rotation, and retained SH coefficients of $\mathcal P_{t-1}$ remain fixed. The new layer is thus optimized in the context in which it is decoded.

Appending Gaussians nevertheless changes transmittance and reweights the contribution of the existing prefix. SHIN reserves opacity as the only mutable prefix parameter. During enhancement refinement,
\begin{equation}
o_i^{(t)}=o_i^{(t-1)}+\Delta o_i^{(t)},
\qquad
\omega_i^{(t)}=\sigma\!\left(o_i^{(t)}\right).
\label{eq:opacity-update}
\end{equation}
Here, $\omega_i^{(t)}$ denotes the activated scalar opacity, distinct from the pixel-dependent compositing term $\alpha_i(x)$ in \cref{eq:alpha-compositing}. At decoding, the opacity deltas are applied to the existing prefix before the new Gaussian records are appended. Each stage therefore improves the decoded representation without retransmitting earlier geometry or SH coefficients.

\paragraph{Loss roles.}
The Laplacian-band $\mathcal L_1$ loss and clamped RGB MSE are one-time probes for Gaussian ordering and Layer--Degree Allocation, respectively; only the stage-wise objectives optimize transmitted parameters.

%% file: sec/3_perlayer_float.tex
\begin{figure}[!t]
\centering
\includegraphics[width=\columnwidth]{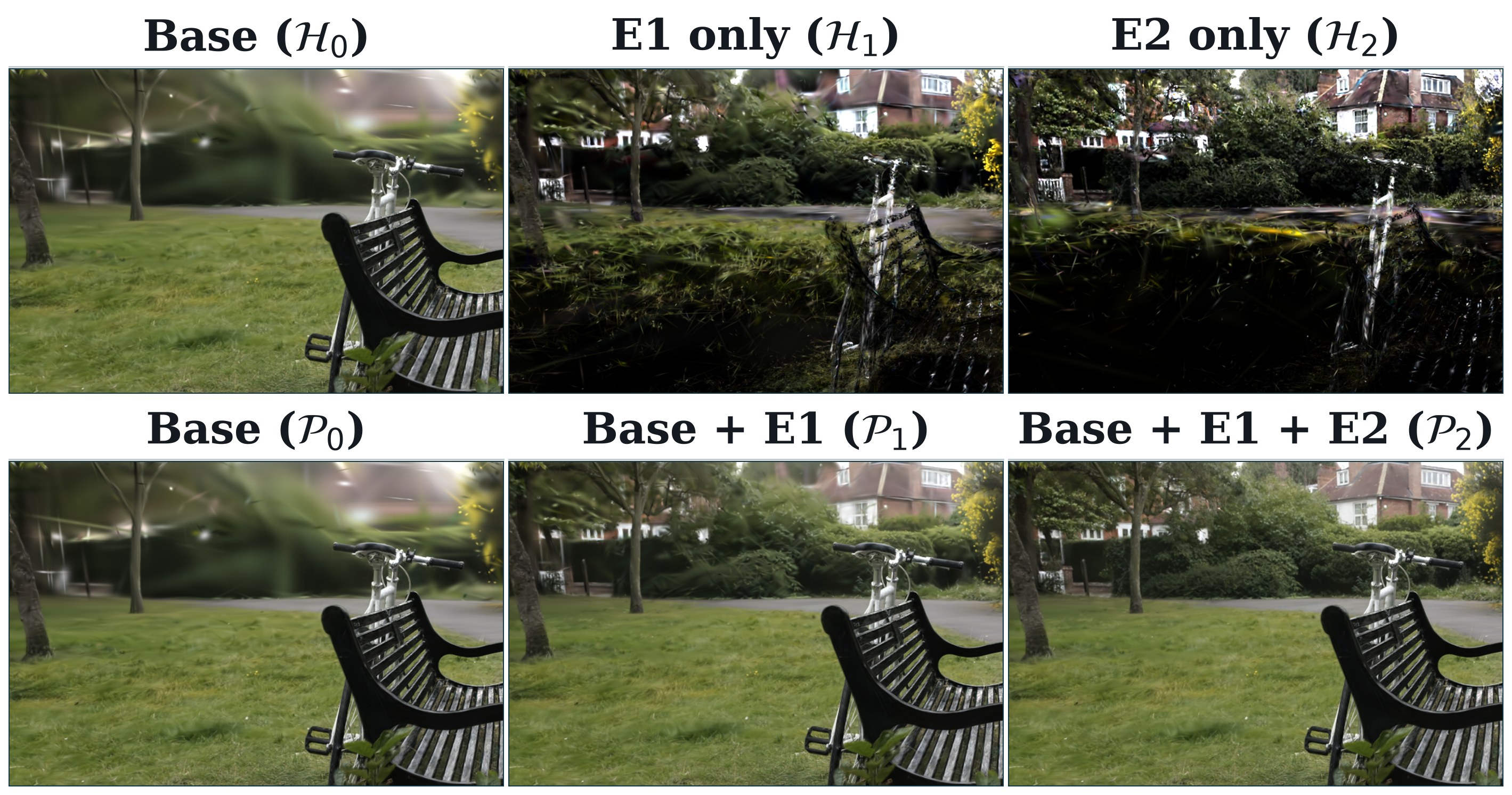}
\caption{\textbf{Disjoint layer contents and cumulative prefixes on Bicycle.} Top: individual layer contents ($\mathcal H_0,\mathcal H_1,\mathcal H_2$) rendered in isolation. Bottom: cumulative prefixes ($\mathcal P_0,\mathcal P_1,\mathcal P_2$).}
\label{fig:per_layer_effect}
\end{figure}

%% file: sec/3_main_results_floats.tex
\begin{figure*}[!t]
\centering
\includegraphics[width=\textwidth]{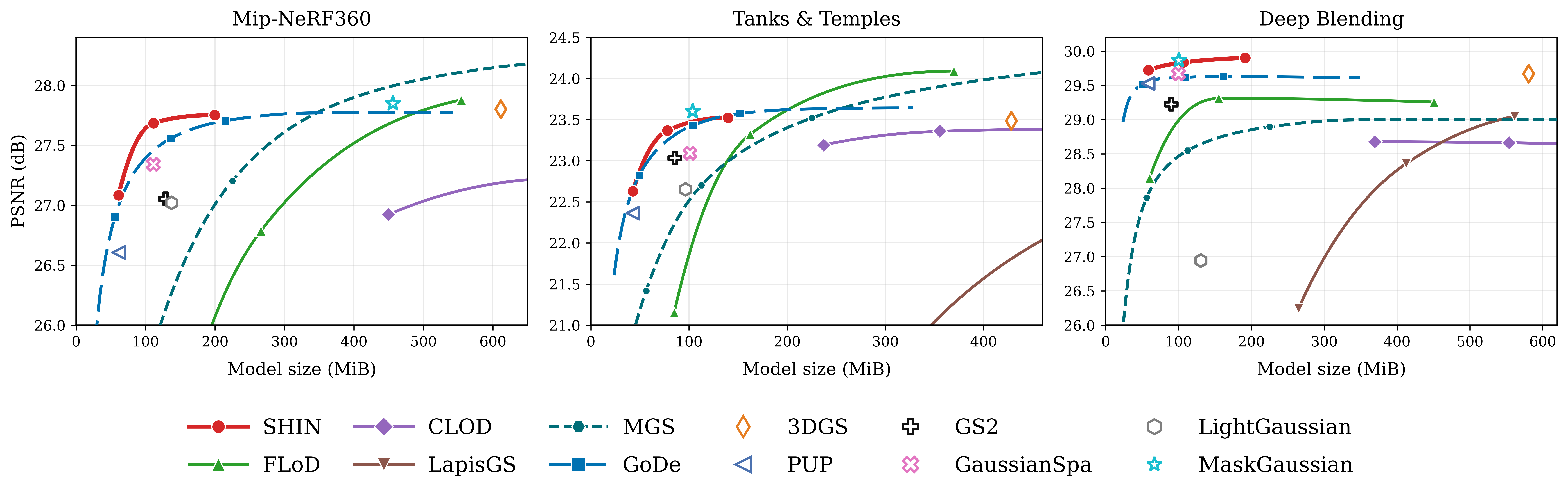}
\caption{\textbf{PSNR--model-size trade-offs.} Curves denote multi-rate representations and markers denote independently optimized compact models; SHIN stages form cumulative prefixes.}
\label{fig:main_rd_psnr}
\end{figure*}

%% file: sec/3_main_results_table.tex
\begin{table*}[!t]
\centering
\caption{Quality, model-size, and rendering-speed comparison. Size denotes cumulative model size; parentheses show the additional size from the preceding operating point for SHIN, MGS, and LapisGS. Other methods report total model size at each operating point.}
\label{tab:main_progressive_lod_3point}
\scriptsize
\setlength{\tabcolsep}{1.45pt}
\renewcommand{\arraystretch}{1.08}
\begin{tabular*}{\textwidth}{@{\extracolsep{\fill}}ll*{3}{ccccc}@{}}
\toprule
\multirow{2}{*}{Stage} & \multirow{2}{*}{Method}
& \multicolumn{5}{c}{Mip-NeRF360}
& \multicolumn{5}{c}{Tanks\&Temples}
& \multicolumn{5}{c}{Deep Blending} \\
\cmidrule(lr){3-7}\cmidrule(lr){8-12}\cmidrule(lr){13-17}
& & PSNR$\uparrow$ & SSIM$\uparrow$ & LPIPS$\downarrow$ & \makecell{Size\\(MiB)$\downarrow$} & FPS$\uparrow$
& PSNR$\uparrow$ & SSIM$\uparrow$ & LPIPS$\downarrow$ & \makecell{Size\\(MiB)$\downarrow$} & FPS$\uparrow$
& PSNR$\uparrow$ & SSIM$\uparrow$ & LPIPS$\downarrow$ & \makecell{Size\\(MiB)$\downarrow$} & FPS$\uparrow$ \\
\midrule
Base & \textbf{SHIN}
& 27.08 & .8032 & .2396 & 61 & 326
& 22.63 & .7984 & .2807 & 43 & 219
& 29.72 & .9029 & .2491 & 59 & 448 \\
& MGS~\cite{guo2026matryoshka}
& 24.24 & .6531 & .3839 & 56 & 529
& 21.42 & .7371 & .3661 & 56 & 394
& 27.86 & .8788 & .2906 & 56 & 602 \\
& GoDe~\cite{di2025gode}
& 26.90 & .7842 & .2738 & 56 & 369
& 22.82 & .7868 & .3068 & 49 & 195
& 29.52 & .8950 & .2757 & 51 & 497 \\
& FLoD~\cite{seo2025flod}
& 24.22 & .6389 & .4358 & 119 & 444
& 21.15 & .7085 & .4326 & 85 & 292
& 28.15 & .8689 & .3326 & 60 & 661 \\
& CLoD-GS~\cite{cheng2026clodgs}
& 26.92 & .7984 & .1998 & 450 & 198
& 23.19 & .8114 & .2256 & 237 & 234
& 28.68 & .8960 & .1882 & 369 & 260 \\
& LapisGS~\cite{shi2025lapisgs}
& 21.31 & .5567 & .4335 & 578 & 123
& 19.81 & .6868 & .3836 & 269 & 109
& 26.25 & .8191 & .3214 & 265 & 199 \\
\midrule
+E1 & \textbf{SHIN}
& 27.68 & .8173 & .2169 & 111 (+50) & 216
& 23.37 & .8133 & .2596 & 78 (+35) & 157
& 29.84 & .9035 & .2446 & 107 (+48) & 259 \\
& MGS
& 25.84 & .7252 & .3063 & 113 (+56) & 378
& 22.70 & .7873 & .3024 & 113 (+56) & 279
& 28.55 & .8945 & .2603 & 113 (+56) & 456 \\
& GoDe
& 27.56 & .8149 & .2267 & 136 & 263
& 23.43 & .8080 & .2743 & 104 & 151
& 29.62 & .8983 & .2613 & 110 & 406 \\
& FLoD
& 26.79 & .7674 & .2899 & 266 & 283
& 23.32 & .7955 & .3089 & 162 & 216
& 29.31 & .8971 & .2737 & 155 & 423 \\
& CLoD-GS
& 27.22 & .8097 & .1830 & 675 & 142
& 23.36 & .8193 & .2106 & 356 & 176
& 28.66 & .8970 & .1830 & 554 & 189 \\
& LapisGS
& 24.87 & .6996 & .3182 & 1020 (+442) & 76
& 22.22 & .7633 & .3166 & 490 (+221) & 71
& 28.36 & .8725 & .2517 & 413 (+148) & 130 \\
\midrule
+E2 & \textbf{SHIN}
& 27.75 & .8202 & .2104 & 200 (+88) & 159
& 23.52 & .8182 & .2514 & 140 (+62) & 122
& 29.90 & .9038 & .2427 & 192 (+85) & 166 \\
& MGS
& 27.20 & .7931 & .2295 & 225 (+113) & 257
& 23.52 & .8242 & .2449 & 225 (+113) & 187
& 28.90 & .9040 & .2391 & 225 (+113) & 319 \\
& GoDe
& 27.70 & .8227 & .2116 & 215 & 211
& 23.58 & .8142 & .2627 & 152 & 132
& 29.63 & .8988 & .2566 & 161 & 348 \\
& FLoD
& 27.88 & .8186 & .2204 & 554 & 160
& 24.09 & .8275 & .2517 & 370 & 124
& 29.26 & .9025 & .2507 & 451 & 191 \\
& CLoD-GS
& 27.25 & .8129 & .1772 & 900 & 110
& 23.38 & .8218 & .2055 & 474 & 141
& 28.61 & .8967 & .1823 & 739 & 147 \\
& LapisGS
& 26.54 & .7548 & .2702 & 1417 (+397) & 57
& 23.16 & .7911 & .2878 & 758 (+268) & 52
& 29.05 & .8900 & .2272 & 561 (+149) & 101 \\
\midrule
\multirow{6}{*}{Independent}
& 3DGS~\cite{kerbl3Dgaussians}
& 27.80 & .8264 & .2015 & 611 & 141
& 23.49 & .8308 & .2350 & 428 & 110
& 29.67 & .9025 & .2413 & 581 & 145 \\
& PUP~\cite{hanson2025pup}
& 26.61 & .7883 & .2678 & 61 & 719
& 22.36 & .7916 & .2995 & 43 & 419
& 29.53 & .9034 & .2584 & 58 & 765 \\
& GS$^2$~\cite{yang2026gs2}
& 27.05 & .8075 & .2420 & 129 & 213
& 23.03 & .8071 & .2813 & 85 & 107
& 29.22 & .9023 & .2638 & 90 & 231 \\
& GaussianSpa~\cite{zhang2025gaussianspa}
& 27.34 & .8126 & .2378 & 111 & 576
& 23.09 & .8255 & .2408 & 101 & 427
& 29.67 & .9058 & .2535 & 100 & 614 \\
& LightGaussian~\cite{fan2024lightgaussian}
& 27.02 & .8064 & .2368 & 137 & 238
& 22.65 & .7946 & .2970 & 96 & 111
& 26.94 & .8709 & .3084 & 131 & 193 \\
& MaskGaussian~\cite{liu2025maskgaussian}
& 27.85 & .8257 & .2028 & 456 & 182
& 23.60 & .8200 & .2619 & 104 & 238
& 29.87 & .9059 & .2526 & 100 & 541 \\
\bottomrule
\end{tabular*}
\end{table*}

%% file: sec/3_rate_transition_tables.tex
\begin{table*}[!t]
\begin{minipage}[t]{0.485\textwidth}
\vspace{0pt}
\input{sec/3_bdrate_table}
\end{minipage}
\hfill
\begin{minipage}[t]{0.485\textwidth}
\vspace{0pt}
\input{sec/3_transition_table}
\end{minipage}
\end{table*}

%% file: sec/3_bdrate_table.tex
\centering
\captionsetup{skip=2pt}
\caption{Bj{\o}ntegaard metrics relative to SHIN using all post-Pareto points. Positive BD-rate and negative BD-PSNR favor SHIN; dashes denote no overlap. Post-Pareto integration domains are reported in the supplement.}
\label{tab:main_bd_metrics}
\scriptsize
\setlength{\tabcolsep}{1.1pt}
\renewcommand{\arraystretch}{0.92}
\begin{adjustbox}{max width=\linewidth}
\begin{tabular}{@{}l*{3}{cc}@{}}
\toprule
& \multicolumn{2}{c}{Mip-NeRF360}
& \multicolumn{2}{c}{Tanks\&Temples}
& \multicolumn{2}{c}{Deep Blending} \\
\cmidrule(lr){2-3}\cmidrule(lr){4-5}\cmidrule(lr){6-7}
Method & \makecell{BD-rate\\$\downarrow$} & \makecell{BD-PSNR\\$\uparrow$}
& \makecell{BD-rate\\$\downarrow$} & \makecell{BD-PSNR\\$\uparrow$}
& \makecell{BD-rate\\$\downarrow$} & \makecell{BD-PSNR\\$\uparrow$} \\
\midrule
\textbf{SHIN} & 0.0\% & 0.00 & 0.0\% & 0.00 & 0.0\% & 0.00 \\
MGS~\cite{guo2026matryoshka} & +232.2\% & $-1.81$ & +157.3\% & $-1.25$ & -- & $-1.36$ \\
GoDe~\cite{di2025gode} & +41.7\% & $-0.17$ & +12.0\% & $-0.06$ & -- & $-0.21$ \\
FLoD~\cite{seo2025flod} & +397.8\% & $-3.33$ & +150.7\% & $-2.08$ & -- & $-1.09$ \\
CLoD-GS~\cite{cheng2026clodgs} & +882.5\% & $-4.48$ & +266.6\% & $-2.64$ & -- & $-1.57$ \\
LapisGS~\cite{shi2025lapisgs} & -- & -- & +1286.6\% & $-6.20$ & -- & $-6.04$ \\
\bottomrule
\end{tabular}
\end{adjustbox}

%% file: sec/3_transition_table.tex
\centering
\captionsetup{skip=2pt}
\caption{Transition cost on Mip-NeRF360 under incremental delivery. $\Delta$ Data denotes newly required data; Local measures decoder-side update latency, and Total includes transfer at 1~Gbps. Parentheses indicate opacity-update latency.}
\label{tab:transition}
\scriptsize
\setlength{\tabcolsep}{1.5pt}
\renewcommand{\arraystretch}{0.95}
\begin{adjustbox}{max width=\linewidth}
\begin{tabular}{@{}lccc ccc@{}}
\toprule
\multirow{2}{*}{Method}
& \multicolumn{3}{c}{Operating Point 1 $\rightarrow$ 2}
& \multicolumn{3}{c}{Operating Point 2 $\rightarrow$ 3} \\
\cmidrule(lr){2-4}\cmidrule(lr){5-7}
& \makecell{$\Delta$ Data\\MiB$\downarrow$}
& \makecell{Local\\ms$\downarrow$}
& \makecell{Total\\ms$\downarrow$}
& \makecell{$\Delta$ Data\\MiB$\downarrow$}
& \makecell{Local\\ms$\downarrow$}
& \makecell{Total\\ms$\downarrow$} \\
\midrule
\textbf{SHIN}
& \textbf{50} & 9 (0.4) & \textbf{431}
& 88 & 11 (0.7) & 750 \\
MGS~\cite{guo2026matryoshka}
& 56 & \textbf{5} & 477
& 113 & 11 & 955 \\
GoDe~\cite{di2025gode}
& 80 & 7 & 680
& \textbf{78} & \textbf{8} & \textbf{665} \\
FLoD~\cite{seo2025flod}
& 266 & 15 & 2247
& 554 & 30 & 4680 \\
CLoD-GS~\cite{cheng2026clodgs}
& 225 & 24 & 1911
& 225 & 28 & 1914 \\
LapisGS~\cite{shi2025lapisgs}
& 442 & 41 (0.9) & 3746
& 397 & 45 (1.4) & 3376 \\
\bottomrule
\end{tabular}
\end{adjustbox}

%% file: sec/3_allocation_floats.tex
\begin{figure*}[!t]
\centering
\includegraphics[width=\textwidth]{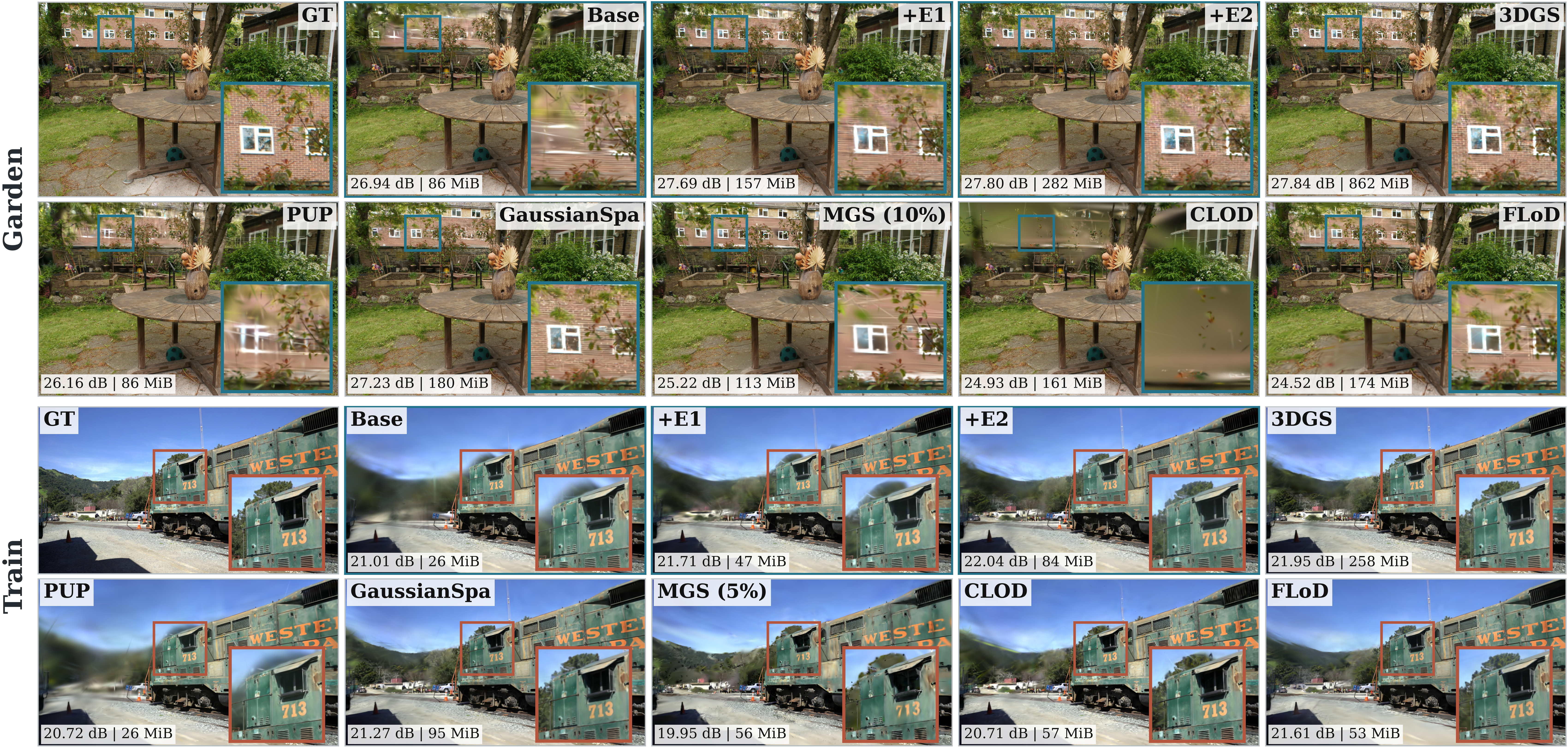}
\caption{\textbf{Qualitative comparison on Garden and Train.} Insets highlight fine structures; labels report PSNR and model size.}
\label{fig:qualitative_progression}
\vspace{0.4em}
\begin{minipage}[t]{0.485\textwidth}
\vspace{0pt}
\centering
\captionsetup{skip=2pt}
\includegraphics[width=0.70\linewidth]{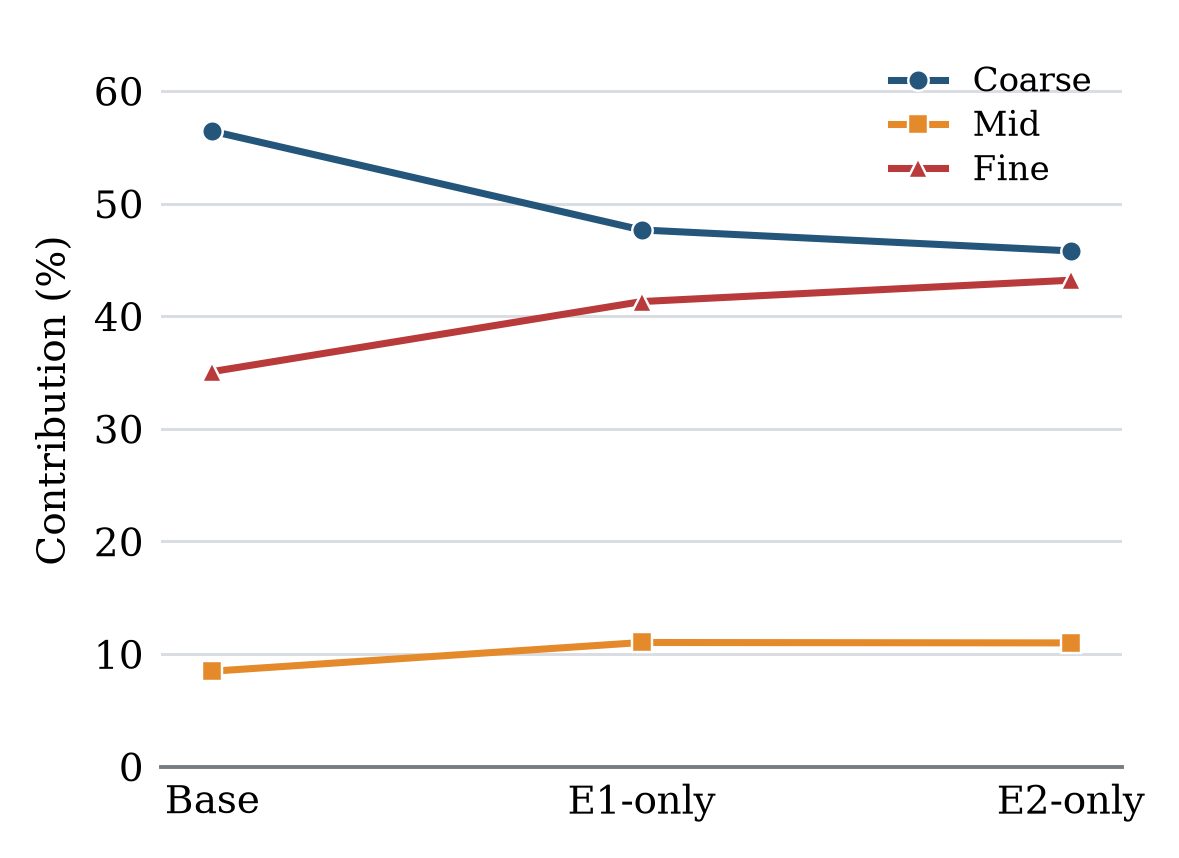}
\caption{\textbf{Scale-wise position--scale sensitivity of newly added Gaussians on Mip-NeRF360.} E1/E2 denote individual layers.}
\label{fig:multiscale_profile}
\end{minipage}
\hfill
\begin{minipage}[t]{0.485\textwidth}
\vspace{0pt}
\centering
\captionsetup{skip=2pt}
\includegraphics[width=0.90\linewidth]{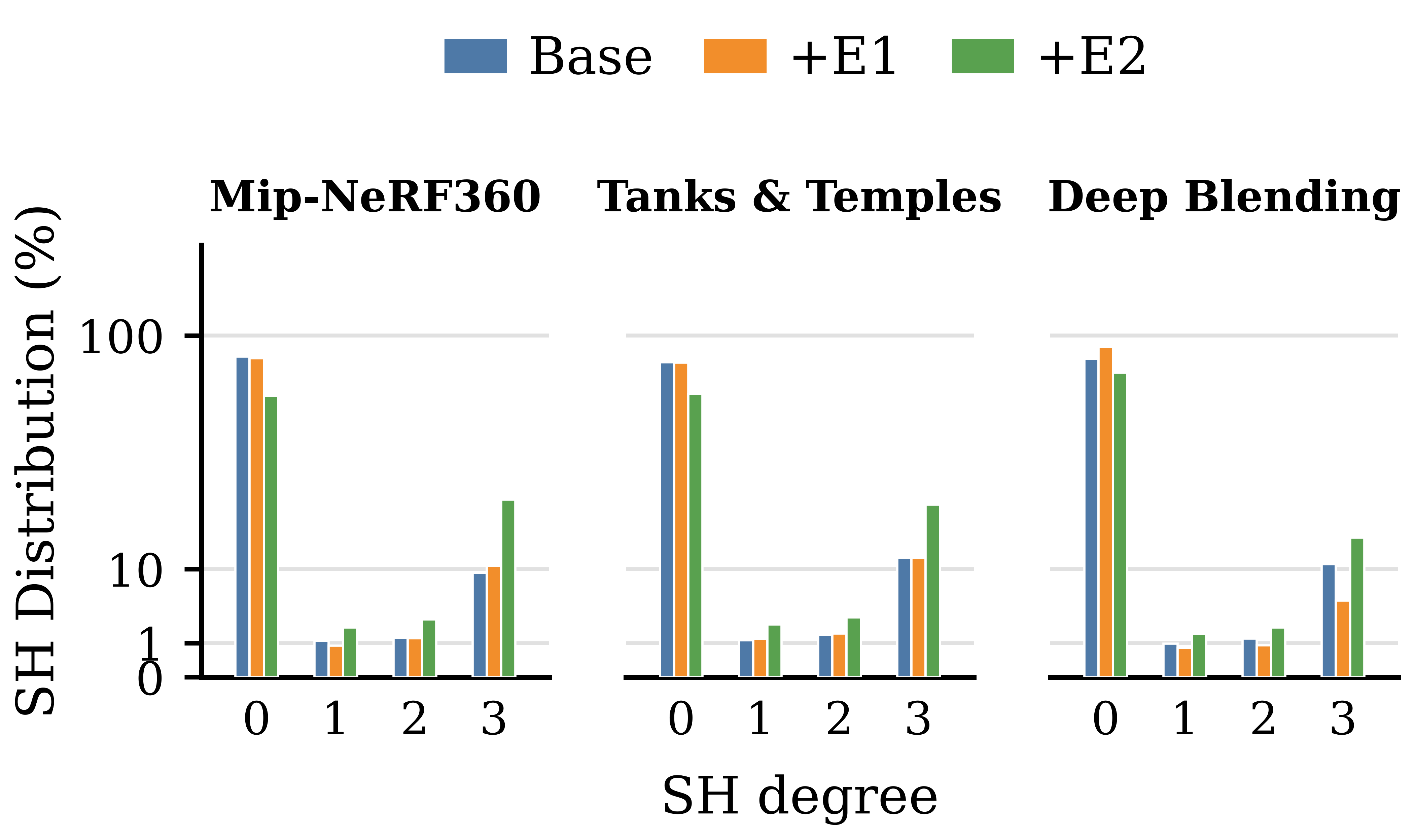}
\caption{\textbf{Layer-wise SH-degree distributions across datasets.}}
\label{fig:sh_distribution}
\end{minipage}
\vspace{-1.0em}
\end{figure*}

%% file: sec/3_experimental_results.tex
\section{Experimental Results}
\label{sec:experimental_results}

\noindent\textbf{Datasets.}
Evaluation follows the standard 3DGS benchmark protocol on Mip-NeRF360 \cite{barron2022mipnerf360}, Tanks\&Temples \cite{Knapitsch2017}, and Deep Blending \cite{DeepBlending2018}.

\noindent\textbf{Implementation.}
SHIN starts from 3DGS models \cite{kerbl3Dgaussians} trained for 30k iterations. The cumulative Gaussian payloads at the three operating points---Base, Base+E1, and Base+E1+E2---are set to 10\%, 18\%, and 32\% of the source-model Gaussian payload, respectively. Thus, E1 and E2 contain only the incremental Gaussian records required to move between successive cumulative targets. Each stage is refined for 5k iterations without densification. Reported model sizes additionally include the opacity-update data. All reported sizes are measured from the serialized artifacts used for rendering and decoding. Quality is evaluated with PSNR, SSIM \cite{wang2004ssim}, and LPIPS \cite{zhang2018lpips}. Rendering speed (FPS) is measured on an RTX 6000 Ada.

\noindent\textbf{Comparisons.}
Comparisons cover five multi-rate and progressive 3DGS
methods: MGS, FLoD, CLoD-GS, GoDe, and LapisGS.
Table~1 reports three representative operating points for
each method, with all model sizes measured from the
corresponding PLY representations under a common
storage-accounting protocol. Independent compact
references include 3DGS, PUP, GS$^2$, GaussianSpa,
LightGaussian, and MaskGaussian. PSNR BD-rate and
BD-PSNR~\cite{bjontegaard2001} are computed from all
measured Pareto-filtered dataset-average curves using
PCHIP over their exact common PSNR and log-model-size
intervals, respectively, without extrapolation
(Supp.~Tab.~\ref{tab:bdrate_common_psnr_intervals}).
Transition measurements follow consecutive operating
points and count only newly required data.

\noindent\textbf{Cumulative quality and rate efficiency.}

\Cref{fig:main_rd_psnr,tab:main_progressive_lod_3point}
evaluate SHIN as a sequence of reusable operating points.
Later layers improve local structure and view-dependent
appearance while retaining the geometry and SH records
decoded at the preceding point and appending only the
corresponding enhancement payload. The resulting gains
therefore extend one shared representation rather than
replace it with independently optimized models.

\Cref{tab:main_bd_metrics} provides matched-quality and
matched-size summaries. SHIN has lower average model size
over every common PSNR interval and higher average PSNR
over every common model-size interval, including Deep
Blending where the measured PSNR ranges do not overlap.

\noindent\textbf{Transition cost under cumulative reuse.}
\Cref{tab:transition} reports Mip-NeRF360 at 1~Gbps. SHIN adds only 50 and 88~MiB for the two transitions without replacing geometry or SH records. Across three datasets and 100~Mbps--1~Gbps, it ranks within the top two in total latency for every transition and first for all Base-to-E1 transitions (Supp.~Tab.~\ref{tab:transition_all_datasets_all_bandwidths}). Applying the opacity deltas takes 0.4 and 0.7~ms.

\noindent\textbf{Independently optimized compact models.}
The lower block of \cref{tab:main_progressive_lod_3point} provides fixed-budget quality references. At comparable storage, SHIN remains competitive with independently optimized compact models while providing three reusable operating points from one cumulative asset.

\input{sec/3_ablation_figure}

\noindent\textbf{Layer-wise progression and qualitative comparison.}
\Cref{fig:per_layer_effect,fig:qualitative_progression} show
that E1 and E2 add complementary spatial and appearance
detail, yielding higher PSNR at similar storage or comparable
PSNR at substantially lower storage while preserving the
global scene layout.

\subsection{Allocation Analysis}
\label{sec:allocation_analysis}

\noindent\textbf{Structural progression along the shared order.}
\Cref{fig:multiscale_profile} decomposes the position--scale sensitivity of newly added Gaussians by pyramid band. From Base to E2, the coarse share decreases from 56.4\% to 45.8\% while the fine share rises from 35.1\% to 43.2\%, showing mixed scales with increasing emphasis on finer correction.

\noindent\textbf{Degree-allocation pattern.}
\Cref{fig:sh_distribution} shows that Layer--Degree Allocation
concentrates on SH0 and SH3, with higher degrees increasingly
common in later layers and directing capacity toward richer
view-dependent appearance.

%% file: sec/3_ablation_figure.tex
\begin{table*}[!t]
\centering
\captionsetup{skip=2pt}
\caption{Ablation on Mip-NeRF360. Sizes include opacity-update data where applicable; BD-rate uses SHIN as anchor.}
\label{tab:mip_unified_ablation}
\scriptsize
\setlength{\tabcolsep}{1.8pt}
\renewcommand{\arraystretch}{0.90}
\begin{adjustbox}{max width=\textwidth}
\begin{tabular}{@{}ll*{3}{cccc}c@{}}
\toprule
\multirow{2}{*}{Component}
& \multirow{2}{*}{Variant}
& \multicolumn{4}{c}{Base}
& \multicolumn{4}{c}{+E1}
& \multicolumn{4}{c}{+E2}
& \multirow{2}{*}{BD-rate$\downarrow$} \\
\cmidrule(lr){3-6}\cmidrule(lr){7-10}\cmidrule(lr){11-14}
& & PSNR$\uparrow$ & SSIM$\uparrow$ & LPIPS$\downarrow$ & MiB$\downarrow$
& PSNR$\uparrow$ & SSIM$\uparrow$ & LPIPS$\downarrow$ & MiB$\downarrow$
& PSNR$\uparrow$ & SSIM$\uparrow$ & LPIPS$\downarrow$ & MiB$\downarrow$ & \\
\midrule
Ordering & PUP-style position--scale order~\cite{hanson2025pup} & 26.89 & .8079 & .2280 & 61 & 27.50 & .8147 & .2168 & 111 & 27.59 & .8164 & .2139 & 200 & +28.6\% \\
\midrule
\multirow{4}{*}{Degree assignment} & Fixed SH0 & 26.78 & .8053 & .2314 & 61 & 27.15 & .8119 & .2191 & 111 & 27.16 & .8122 & .2173 & 173 & +63.9\% \\
 & Fixed SH1 & 26.54 & .7958 & .2497 & 61 & 27.20 & .8113 & .2251 & 111 & 27.35 & .8170 & .2138 & 200 & +94.4\% \\
 & Fixed SH2 & 25.91 & .7751 & .2806 & 61 & 27.02 & .8016 & .2434 & 111 & 27.51 & .8171 & .2187 & 200 & +123.6\% \\
 & Fixed SH3 & 25.13 & .7530 & .3116 & 61 & 26.46 & .7815 & .2733 & 111 & 27.39 & .8085 & .2348 & 200 & +178.4\% \\
\midrule
\multirow{2}{*}{Allocation} & Sequential Gaussian-first & 26.78 & .8053 & .2314 & 61 & 27.15 & .8119 & .2191 & 111 & 27.23 & .8128 & .2166 & 200 & +100.1\% \\
 & Endpoint-first truncation & 26.92 & .7999 & .2454 & 61 & 27.60 & .8172 & .2178 & 111 & 27.67 & .8204 & .2105 & 200 & +14.3\% \\
\midrule
Calibration & w/o score calibration & 26.80 & .8054 & .2313 & 61 & 27.17 & .8120 & .2190 & 111 & 27.23 & .8127 & .2166 & 200 & +88.7\% \\
\midrule
Refinement & w/o opacity update & 27.08 & .8032 & .2396 & 61 & 27.53 & .8151 & .2200 & 110 & 27.59 & .8175 & .2154 & 196 & +14.4\% \\
\midrule
Full & \textbf{SHIN} & \textbf{27.08} & \textbf{.8032} & \textbf{.2396} & \textbf{61} & \textbf{27.68} & \textbf{.8173} & \textbf{.2169} & \textbf{111} & \textbf{27.75} & \textbf{.8202} & \textbf{.2104} & \textbf{200} & \textbf{0.0\%} \\
\bottomrule
\end{tabular}
\end{adjustbox}
\vspace{-1.0em}
\end{table*}

%% file: sec/3_ablation.tex
\section{Ablation Study}
\label{sec:ablation}

The ablations isolate Gaussian ordering, degree assignment,
score calibration, and opacity adaptation while keeping the
cumulative budgets and all other settings fixed.

\par\addvspace{0.3\baselineskip}
\noindent\textbf{Gaussian ordering.}
Replacing the multi-scale order with the PUP-style single-scale order raises BD-rate by 28.6\%, with allocation and refinement unchanged. The gap isolates the cost of reusing one scale across short and long prefixes: multi-scale aggregation supports broad Base coverage and finer later residuals.

\par\addvspace{0.3\baselineskip}
\noindent\textbf{Degree assignment.}
Fixed degrees exhibit opposite failure modes: higher degrees reduce Base spatial coverage, whereas lower degrees cannot enrich appearance later. Accordingly, the best fixed degree changes from SH0 at Base to SH1 at +E1 and SH2 at +E2; no single degree remains efficient along the full trajectory.

With fixed SH0, all source Gaussians have been added by +E2, so no feasible spatial or appearance action remains; the +E1-to-+E2 gain collapses to 0.01~dB.

\par\addvspace{0.3\baselineskip}
\noindent\textbf{Joint allocation.}
Gaussian-first commits bytes to spatial support before richer appearance and incurs a 100.1\% BD-rate penalty. Endpoint-first degrades earlier prefixes by 14.3\%. Thus, Gaussian addition and retained SH degrees must be decided together as the hierarchy grows.

\par\addvspace{0.3\baselineskip}
\noindent\textbf{Score calibration.}
The structural and appearance scores originate from different probes and are not directly comparable. Removing calibration lowers PSNR by 0.29--0.52~dB, whereas calibrated variants from the median to Q99 remain closely matched (Supp.~Sec.~\ref{sec:supp-calibration-percentile}). Thus, scale alignment matters more than the particular robust statistic.

\par\addvspace{0.3\baselineskip}
\noindent\textbf{Opacity update.}
Appending Gaussians changes transmittance and reweights the existing prefix. Without opacity updates, this mismatch raises BD-rate by 14.4\%, while the deltas add only 1.2--2.0\% to cumulative size and under 1~ms per transition. Appending previously omitted higher-order SH bands adds only 0.11~dB at +E2 while increasing cumulative size to about twice that of SHIN on Mip-NeRF360 (Supp.~Sec.~\ref{sec:supp-sh-bands}). Opacity therefore adapts compositing at low cost while keeping transmitted spatial and SH parameters reusable.

%% file: sec/4_conclusion.tex
\section{Conclusion}
\label{sec:conclusion}

A scalable 3DGS representation must improve quality while reusing the representation already delivered. SHIN combines a shared multi-scale Gaussian order, Layer--Degree Allocation that jointly decides which Gaussians are added and how much SH representation they retain, and lightweight opacity updates that adapt compositing without retransmitting geometry or SH coefficients. Successive enhancements improve a single cumulative representation using only incremental enhancement payloads, comprising newly added Gaussian records and opacity deltas for the resident prefix. SHIN also achieves the lowest average rate in every overlapping comparison with the multi-rate and progressive methods. Allocation analysis and ablations show that early capacity establishes broad spatial support, later capacity refines local structure and view-dependent appearance, and fixed degrees or separate spatial and appearance decisions weaken this progression. These results establish joint spatial--appearance allocation as central to scalable 3DGS.

%% file: sec/supplementary_implementation.tex
\section{Byte Accounting}
\label{sec:supp-hierarchy}

\subsection{Gaussian Record Cost}

Because \cref{alg:layer_degree} appends Gaussians in the shared order, each enhancement contains a contiguous suffix of that order.

The allocation model follows the float32 PLY layout. Each record contains position (12 bytes), three normal placeholders retained by the PLY layout (12 bytes), DC color (12 bytes), opacity (4 bytes), scale (12 bytes), rotation (16 bytes), and retained higher-order SH coefficients. Consequently,
\begin{equation}
\begin{aligned}
b(d)&=68+12\bigl((d+1)^2-1\bigr),\\
b(0{:}3)&=(68,104,164,248)\ \text{bytes}.
\end{aligned}
\label{eq:supp-record-cost}
\end{equation}

\paragraph{Mixed-degree PLY serialization.}
After hierarchy construction and all layer assignments are complete, sorting is applied only during serialization: within each layer, Gaussian records are sorted by retained SH degree and stored as up to four degree-homogeneous binary PLY blocks (SH0--SH3). Each block has a fixed property schema and records its vertex count in the PLY header, so the decoder obtains $d_i$ from the block being parsed without a per-Gaussian degree tag or index. Reported Gaussian payloads sum the block record bodies; the corresponding PLY headers add less than 0.004~MiB per layer and do not alter any displayed model-size value.

\subsection{Stored Size versus Transition Payload}

For SHIN, \cref{tab:main_progressive_lod_3point} reports each cumulative stored footprint followed by the incremental size from the preceding point in parentheses. Transition $\Delta$ Data is computed from the underlying unrounded cumulative sizes, so it may differ slightly from subtracting the rounded values displayed in the table.

\input{sec/supp_bdrate_intervals_table}
\input{sec/supp_ssim_bd_table}

\section{Score Probes}
\label{sec:supp-ordering}
\label{sec:supp-allocation}

\subsection{Multi-Scale Structural Decomposition}

\paragraph{Pyramid bands.}
All experiments use the same fixed three-level dyadic Laplacian pyramid at scales $1,2,4$, where $\mathcal B_4$ is the coarse low-pass component and $\mathcal B_2$ and $\mathcal B_1$ denote the intermediate and fine residual bands, respectively. The same construction is used for every scene and dataset.

The structural score in \cref{eq:spatial-sensitivity} decomposes by pyramid scale. Define
\begin{equation}
P_{i,m}=
\sum_v\left(
\left\|\frac{\partial\mathcal L_{v,m}^{\mathrm{lap}}}
{\partial\boldsymbol\mu_i}\right\|_2^2+
\left\|\frac{\partial\mathcal L_{v,m}^{\mathrm{lap}}}
{\partial\mathbf s_i}\right\|_2^2
\right),
\label{eq:supp-scale-score}
\end{equation}
so that
\begin{equation}
P_i=\sum_m P_{i,m}.
\label{eq:supp-scale-decomposition}
\end{equation}
The individual $P_{i,m}$ terms are retained for the scale-wise analysis in \cref{fig:multiscale_profile}. This decomposition exposes the evidence contributing to the shared order without assigning a pyramid band to a hierarchy stage.

\subsection{Gate Probes and Score Calibration}

The SH score in \cref{eq:degree-utility} is computed by accumulating squared SH-band gate gradients over the training views. For each view, the gates are initialized to one and one backward pass evaluates the clamped RGB mean-squared error. Squared gradients are accumulated only for Gaussians with positive projected radius in that view. The opacity reference gate is inserted after sigmoid activation and before rasterization:
\begin{equation}
\omega_i^{(z)}=z_i\sigma(o_i),
\qquad z_i=1.
\label{eq:supp-opacity-probe}
\end{equation}
Its view-accumulated sensitivity is
\begin{equation}
G_i^{\mathrm{op}}=
\sum_{v:i\in\mathcal I_v}
\left(
\left.\frac{\partial\mathcal E_v}{\partial z_i}
\right|_{z_i=1}
\right)^2,
\qquad
\mathcal I_v=\{i:r_i^{(v)}>0\}.
\label{eq:supp-opacity-score}
\end{equation}
Finite positive entries are used for scene-level calibration:
\begin{equation}
S_i=\gamma P_i,
\qquad
\gamma=
\frac{Q_{0.95}(G^{\mathrm{op}}_{>0})}
{\max\!\left(Q_{0.95}(P_{>0}),10^{-12}\right)}.
\label{eq:supp-calibration}
\end{equation}
If either positive-entry set is empty, all calibrated structural scores are set to zero. The gates are temporary, untransmitted probes, and their scores are reused across all stages. Because the opacity gate is evaluated in the same clamped image-loss and multiplicative-gate domain as the SH probes, it provides a scene-level reference for score scaling while including the effect of alpha compositing.

\section{Opacity-Delta Serialization and Decoding}
\label{sec:supp-refinement}

Each opacity update stores one IEEE-754 binary16 pre-sigmoid logit delta for every Gaussian in the resident prefix. Records follow the deterministic resident decoding order established by the serialized degree blocks, so no Gaussian indices are required. The raw update payload and cumulative raw payload over $T$ enhancements are therefore
\begin{equation}
R_t^{\mathrm{op,raw}}=2|\mathcal P_{t-1}|,
\qquad
R_{\mathrm{total}}^{\mathrm{op,raw}}
=2\sum_{t=1}^{T}|\mathcal P_{t-1}|
\quad\text{bytes}.
\label{eq:supp-total-opacity-cost}
\end{equation}
At decoding, the deltas are added to the resident opacity logits before the new Gaussian layer is appended.

\section{Evaluation Protocol and Additional Results}
\label{sec:supp-experiments}

\subsection{Bj{\o}ntegaard Integration Intervals}

The Base, +E1, and +E2 comparisons use the method-specific operating points listed in \cref{sec:experimental_results}. \Cref{tab:bdrate_common_psnr_intervals} reports the post-Pareto common PSNR and model-size intervals used for BD-rate and BD-PSNR, respectively. Both metrics use the same unrounded dataset-average curves and omit pairs without overlap in the corresponding integration domain. \cref{tab:ssim_bd_table1_vs_shin} gives secondary SSIM results from the three Table~1 points.

\subsection{Representative Construction Cost}

A representative single-run profile for Garden on an RTX 6000 Ada is reported in \cref{tab:supp-construction-cost}. Refinement accounts for 85.5\% of the 891.3-s total, with ordering, SH scoring, and Layer--Degree Allocation accounting for the remaining 14.5\%; the profile is not an average across scenes.

\begin{table}[H]
\centering
\caption{Construction profile for Garden on an RTX 6000 Ada.}
\label{tab:supp-construction-cost}
\scriptsize
\setlength{\tabcolsep}{5.0pt}
\begin{tabular}{lr}
\toprule
Measurement & Value \\
\midrule
Multi-scale ordering & 14.2 s \\
SH-band scoring & 10.5 s \\
Layer--Degree Allocation & 104.8 s \\
Base refinement & 139.3 s \\
E1 refinement & 263.1 s \\
E2 refinement & 359.4 s \\
\midrule
Total & 891.3 s \\
Peak GPU memory & 11.91 GiB \\
\bottomrule
\end{tabular}
\end{table}

\input{sec/supp_transition_bandwidth_table}

\subsection{Bandwidth Sensitivity of Transition Cost}

Given the measured local latency and serialized transition payload, total latency at bandwidth $B$ is
\begin{equation}
T_{\mathrm{total}}^{\mathrm{ms}}(B)
=
T_{\mathrm{local}}^{\mathrm{ms}}
+
10^3\frac{8\cdot2^{20}\Delta S_{\mathrm{MiB}}}{B},
\label{eq:supp-transition-total}
\end{equation}
where $B$ is the link bandwidth in bits per second and $\Delta S_{\mathrm{MiB}}$ is the newly delivered payload in MiB. \Cref{tab:transition_all_datasets_all_bandwidths} reports the resulting latency at 1~Gbps, 500~Mbps, and 100~Mbps.

\subsection{Cost of Appending Omitted Higher-Order SH Bands}
\label{sec:supp-sh-bands}

This diagnostic evaluates an additive alternative that preserves the same Base and Gaussian hierarchy as SHIN but appends previously omitted higher-order SH bands at later stages. Standard opacity updates are retained, and all previously transmitted geometry and SH coefficients remain unchanged. Each added SH record stores a source identifier and the newly enabled coefficients.

\begin{table}[H]
\centering
\caption{Appending omitted SH bands on Mip-NeRF360. Size is cumulative, and BD-rate is relative to opacity-only SHIN.}
\label{tab:supp-sh-update}
\scriptsize
\setlength{\tabcolsep}{2.0pt}
\renewcommand{\arraystretch}{0.98}
\begin{adjustbox}{max width=\columnwidth}
\begin{tabular}{llccccc}
\toprule
Stage & Prefix update & PSNR$\uparrow$ & SSIM$\uparrow$ & LPIPS$\downarrow$ & MiB$\downarrow$ & BD-rate$\downarrow$ \\
\midrule
Base & Shared & 27.08 & .8032 & .2396 & 61 & -- \\
+E1 & Opacity only & 27.68 & .8173 & .2169 & \textbf{111} & -- \\
& Opacity $+$ added SH bands & \textbf{27.79} & \textbf{.8205} & \textbf{.2131} & 228 & -- \\
+E2 & Opacity only & 27.75 & .8202 & .2104 & \textbf{200} & \textbf{0.00\%} \\
& Opacity $+$ added SH bands & \textbf{27.86} & \textbf{.8236} & \textbf{.2062} & 406 & +14.92\% \\
\bottomrule
\end{tabular}
\end{adjustbox}
\end{table}

At +E2, this alternative gains only 0.1109~dB while increasing size from 200 to 406~MiB and worsening BD-rate by 14.92\%. Thus, in this cumulative setting, appending omitted SH bands is a poor rate--distortion trade-off; opacity-only prefix adaptation more efficiently accommodates later Gaussian additions without retransmitting geometry or SH coefficients.

\subsection{Increasing the Number of Stages}

A five-stage hierarchy uses cumulative Gaussian-payload budgets of 10\%, 14\%, 18\%, 25\%, and 32\%. The three- and five-stage schedules use matched cumulative refinement computation at their shared 10\%, 18\%, and 32\% checkpoints.

\begin{table}[H]
\centering
\caption{Three- versus five-stage hierarchy on Mip-NeRF360. Sizes are five-stage cumulative footprints.}
\label{tab:supp-five-stage}
\scriptsize
\setlength{\tabcolsep}{3.2pt}
\begin{tabular}{ccccc}
\toprule
Budget & 3-stage & 5-stage & $\Delta$ PSNR & MiB \\
\midrule
10\% & 27.084 & 27.085 & +0.001 & 61 \\
14\% & -- & 27.503 & -- & 87 \\
18\% & 27.685 & 27.648 & $-0.037$ & 113 \\
25\% & -- & 27.717 & -- & 158 \\
32\% & 27.753 & 27.734 & $-0.019$ & 204 \\
\bottomrule
\end{tabular}
\end{table}

At the shared checkpoints, the schedules differ by at most 0.037~dB, showing that finer granularity preserves reconstruction quality. The added transitions raise cumulative opacity-update payload from 4 to 9~MiB, quantifying its cost.

\section{Calibration Robustness}
\label{sec:supp-ablation}

\subsection{Calibration Percentile}
\label{sec:supp-calibration-percentile}

\begin{table}[H]
\centering
\caption{Calibration-percentile robustness on Mip-NeRF360. PSNR is in dB; Q95 is the default.}
\label{tab:supp_calibration_robustness}
\scriptsize
\setlength{\tabcolsep}{3.0pt}
\begin{tabular}{lcccc}
\toprule
Calibration & Base & +E1 & +E2 & BD-rate$\downarrow$ \\
\midrule
Q50 & 27.09 & 27.67 & 27.75 & +2.18\% \\
Q90 & 27.06 & \textbf{27.68} & \textbf{27.76} & +1.06\% \\
Q95 & 27.08 & \textbf{27.68} & 27.75 & \textbf{0.00\%} \\
Q99 & \textbf{27.16} & 27.67 & 27.71 & +0.08\% \\
\bottomrule
\end{tabular}
\end{table}

Closely matched variants show that score-scale alignment matters more than percentile choice.

%% file: sec/supp_bdrate_intervals_table.tex
\begin{table*}[!t]
\centering
\caption{Post-Pareto intervals for Bj{\o}ntegaard metrics. Counts are SHIN/baseline; BD-rate uses common PSNR and BD-PSNR uses common log-model-size. Displayed PSNR and size bounds are rounded to 0.01~dB and the nearest MiB, respectively; integrations use unrounded values. Dashes denote no overlap.}
\label{tab:bdrate_common_psnr_intervals}
\label{tab:bdpsnr_common_size_intervals}
\scriptsize
\setlength{\tabcolsep}{3.8pt}
\renewcommand{\arraystretch}{1.05}
\begin{adjustbox}{max width=\textwidth}
\begin{tabular}{@{}llccccc@{}}
\toprule
Dataset & Method & \makecell{Post-Pareto pts.\\(SHIN/method)}
& Common PSNR interval & BD-rate
& Common size interval (MiB) & BD-PSNR (dB) \\
\midrule
\multirow{5}{*}{Mip-NeRF360} & MGS & 3/12 & [27.08, 27.75] & +232.2\% & [61, 200] & $-1.81$ \\
 & GoDe & 3/8 & [27.08, 27.75] & +41.7\% & [61, 200] & $-0.17$ \\
 & FLoD & 3/5 & [27.08, 27.75] & +397.8\% & [61, 200] & $-3.33$ \\
 & CLoD-GS & 3/6 & [27.08, 27.25] & +882.5\% & [61, 200] & $-4.48$ \\
 & LapisGS & 3/4 & -- & -- & -- & -- \\
\midrule
\multirow{5}{*}{Tanks\&Temples} & MGS & 3/12 & [22.63, 23.52] & +157.3\% & [43, 140] & $-1.25$ \\
 & GoDe & 3/8 & [22.63, 23.52] & +12.0\% & [43, 140] & $-0.06$ \\
 & FLoD & 3/5 & [22.63, 23.52] & +150.7\% & [43, 140] & $-2.08$ \\
 & CLoD-GS & 3/6 & [22.63, 23.38] & +266.6\% & [43, 140] & $-2.64$ \\
 & LapisGS & 3/4 & [22.63, 23.16] & +1286.6\% & [120, 140] & $-6.20$ \\
\midrule
\multirow{5}{*}{Deep Blending} & MGS & 3/6 & -- & -- & [59, 192] & $-1.36$ \\
 & GoDe & 3/6 & -- & -- & [59, 161] & $-0.21$ \\
 & FLoD & 3/4 & -- & -- & [59, 155] & $-1.09$ \\
 & CLoD-GS & 3/4 & -- & -- & [59, 192] & $-1.57$ \\
 & LapisGS & 3/4 & -- & -- & [140, 192] & $-6.04$ \\
\bottomrule
\end{tabular}
\end{adjustbox}
\end{table*}

%% file: sec/supp_ssim_bd_table.tex
\begin{table*}[!t]
\centering
\caption{SSIM Bj{\o}ntegaard metrics relative to SHIN using the three points in \cref{tab:main_progressive_lod_3point}. Brackets give the common SSIM and size intervals.}
\label{tab:ssim_bd_table1_vs_shin}
\scriptsize
\setlength{\tabcolsep}{2.3pt}
\renewcommand{\arraystretch}{1.05}
\begin{adjustbox}{max width=\textwidth}
\begin{tabular}{@{}l*{3}{cc}@{}}
\toprule
\multirow{2}{*}{Method}
& \multicolumn{2}{c}{Mip-NeRF360}
& \multicolumn{2}{c}{Tanks\&Temples}
& \multicolumn{2}{c}{Deep Blending} \\
\cmidrule(lr){2-3}\cmidrule(lr){4-5}\cmidrule(lr){6-7}
& \makecell{SSIM BD-rate$\downarrow$\\{[SSIM]}}
& \makecell{BD-SSIM$\uparrow$\\{[MiB]}}
& \makecell{SSIM BD-rate$\downarrow$\\{[SSIM]}}
& \makecell{BD-SSIM$\uparrow$\\{[MiB]}}
& \makecell{SSIM BD-rate$\downarrow$\\{[SSIM]}}
& \makecell{BD-SSIM$\uparrow$\\{[MiB]}} \\
\midrule
\textbf{SHIN} & 0.0\% & 0.0000 & 0.0\% & 0.0000 & 0.0\% & 0.0000 \\
MGS & -- & $-0.0927$ [61--200] & +162.4\% [.7984--.8182] & $-0.0437$ [56--140] & +129.7\% [.9029--.9038] & $-0.0108$ [59--192] \\
GoDe & +44.6\% [.8032--.8202] & $-0.0073$ [61--200] & +79.0\% [.7984--.8142] & $-0.0111$ [49--140] & -- & $-0.0056$ [59--161] \\
FLoD & +526.3\% [.8032--.8186] & $-0.1311$ [119--200] & +237.8\% [.7984--.8182] & $-0.0679$ [85--140] & -- & $-0.0156$ [60--192] \\
CLoD-GS & +836.7\% [.8032--.8129] & -- & +189.0\% [.8114--.8182] & -- & -- & -- \\
LapisGS & -- & -- & -- & -- & -- & -- \\
\bottomrule
\end{tabular}
\end{adjustbox}
\end{table*}

%% file: sec/supp_transition_bandwidth_table.tex
\begin{table*}[!t]
\centering
\caption{Transition cost across datasets. Total combines local latency with transfer at 1~Gbps, 500~Mbps, or 100~Mbps; parentheses give opacity-update latency.}
\label{tab:transition_all_datasets_all_bandwidths}
\scriptsize
\setlength{\tabcolsep}{1.5pt}
\renewcommand{\arraystretch}{1.03}
\begin{adjustbox}{max width=\textwidth}
\begin{tabular}{@{}llrrrrrrrrrr@{}}
\toprule
\multirow{2}{*}{Dataset} & \multirow{2}{*}{Method}
& \multicolumn{5}{c}{Operating Point 1 $\rightarrow$ 2}
& \multicolumn{5}{c}{Operating Point 2 $\rightarrow$ 3} \\
\cmidrule(lr){3-7}\cmidrule(lr){8-12}
& & \makecell{$\Delta$ Data\\MiB$\downarrow$}
& \makecell{Local\\ms$\downarrow$}
& \makecell{Total @1G\\ms$\downarrow$}
& \makecell{Total @500M\\ms$\downarrow$}
& \makecell{Total @100M\\ms$\downarrow$}
& \makecell{$\Delta$ Data\\MiB$\downarrow$}
& \makecell{Local\\ms$\downarrow$}
& \makecell{Total @1G\\ms$\downarrow$}
& \makecell{Total @500M\\ms$\downarrow$}
& \makecell{Total @100M\\ms$\downarrow$} \\
\midrule
Mip-NeRF360 & \textbf{SHIN}
& \textbf{50} & 9 (0.4) & \textbf{431} & \textbf{853} & \textbf{4230}
& 88 & 11 (0.7) & 750 & 1489 & 7401 \\
& LapisGS
& 442 & 41 (0.9) & 3746 & 7451 & 37093
& 397 & 45 (1.4) & 3376 & 6707 & 33357 \\
& MGS
& 56 & \textbf{5} & 477 & 949 & 4724
& 113 & 11 & 955 & 1899 & 9451 \\
& GoDe
& 80 & 7 & 680 & 1353 & 6735
& \textbf{78} & \textbf{8} & \textbf{665} & \textbf{1322} & \textbf{6580} \\
& CLoD-GS
& 225 & 24 & 1911 & 3797 & 18889
& 225 & 28 & 1914 & 3801 & 18893 \\
& FLoD
& 266 & 15 & 2247 & 4478 & 22331
& 554 & 30 & 4680 & 9329 & 46524 \\
\midrule
Tanks\&Temples & \textbf{SHIN}
& \textbf{35} & 8 (0.4) & \textbf{303} & \textbf{598} & \textbf{2959}
& 62 & 10 (0.9) & 527 & 1044 & 5179 \\
& LapisGS
& 221 & 23 (0.5) & 1877 & 3731 & 18562
& 268 & 27 (0.9) & 2275 & 4524 & 22509 \\
& MGS
& 56 & \textbf{5} & 477 & 949 & 4725
& 113 & 10 & 955 & 1899 & 9451 \\
& GoDe
& 55 & 8 & 468 & 928 & 4607
& \textbf{48} & \textbf{5} & \textbf{408} & \textbf{811} & \textbf{4033} \\
& CLoD-GS
& 119 & 12 & 1006 & 2000 & 9953
& 119 & 14 & 1008 & 2002 & 9955 \\
& FLoD
& 162 & 9 & 1368 & 2727 & 13599
& 370 & 21 & 3125 & 6228 & 31059 \\
\midrule
Deep Blending & \textbf{SHIN}
& \textbf{48} & 10 (0.5) & \textbf{415} & \textbf{819} & \textbf{4057}
& 85 & 13 (0.9) & 723 & 1433 & 7113 \\
& LapisGS
& 148 & 15 (0.5) & 1256 & 2498 & 12430
& 149 & 17 (0.8) & 1267 & 2517 & 12516 \\
& MGS
& 56 & \textbf{5} & 477 & 949 & 4725
& 113 & 11 & 955 & 1899 & 9451 \\
& GoDe
& 59 & 6 & 499 & 993 & 4944
& \textbf{52} & \textbf{5} & \textbf{437} & \textbf{869} & \textbf{4329} \\
& CLoD-GS
& 185 & 19 & 1569 & 3118 & 15510
& 185 & 22 & 1571 & 3120 & 15512 \\
& FLoD
& 155 & 9 & 1310 & 2610 & 13012
& 451 & 25 & 3808 & 7591 & 37857 \\
\bottomrule
\end{tabular}
\end{adjustbox}
\end{table*}

%% file: main.bib
@inproceedings{barron2022mipnerf360,
  address = {New Orleans, LA, USA},
  author = {Barron, Jonathan T. and Mildenhall, Ben and Verbin, Dor and Srinivasan, Pratul P. and Hedman, Peter},
  booktitle = {Proceedings of the IEEE/CVF Conference on Computer Vision and Pattern Recognition (CVPR)},
  pages = {5470--5479},
  publisher = {IEEE},
  title = {{Mip-NeRF 360: Unbounded Anti-Aliased Neural Radiance Fields}},
  year = {2022}
}

@techreport{bjontegaard2001,
  author = {Bj{\o}ntegaard, Gisle},
  institution = {ITU-T Video Coding Experts Group},
  number = {VCEG-M33},
  title = {{Calculation of average PSNR differences between RD-curves}},
  url = {https://www.itu.int/wftp3/av-arch/video-site/0104_Aus/VCEG-M33.doc},
  year = {2001}
}

@inproceedings{Chen_HAC_ECCV2024,
  author = {Chen, Yihang and Wu, Qianyi and Lin, Weiyao and Harandi, Mehrtash and Cai, Jianfei},
  booktitle = {Computer Vision -- ECCV 2024},
  doi = {10.1007/978-3-031-72667-5_24},
  pages = {422--438},
  title = {{HAC: Hash-Grid Assisted Context for 3D Gaussian Splatting Compression}},
  url = {https://doi.org/10.1007/978-3-031-72667-5_24},
  year = {2025}
}

@inproceedings{Chen_FCGS_ICLR2025,
  author = {Chen, Yihang and Wu, Qianyi and Li, Mengyao and Lin, Weiyao and Harandi, Mehrtash and Cai, Jianfei},
  booktitle = {International Conference on Learning Representations},
  pages = {74859--74872},
  title = {{Fast Feedforward 3D Gaussian Splatting Compression}},
  url = {https://proceedings.iclr.cc/paper_files/paper/2025/file/ba404795c58d122a4b6fc2672d84d9f3-Paper-Conference.pdf},
  volume = {2025},
  year = {2025}
}

@inproceedings{cheng2026clodgs,
  author = {Cheng, Zhigang and Sun, Mingchao and Liu, Yu and Ge, Zengye and Tang, Luyang and Xu, Mu and Li, Yangyan and Pan, Peng},
  booktitle = {International Conference on Learning Representations},
  editor = {C. Vondrick and B. Hariharan and C. Raffel and L. Pinto and D. Yang and A. Faust},
  pages = {89275--89299},
  title = {{CLoD-GS: Continuous Level-of-Detail via 3D Gaussian Splatting}},
  url = {https://proceedings.iclr.cc/paper_files/paper/2026/file/90134bfef0230e6ec7afa0391396490b-Paper-Conference.pdf},
  volume = {2026},
  year = {2026}
}

@article{di2025gode,
  archiveprefix = {arXiv},
  author = {Di Sario, Francesco and Renzulli, Riccardo and Grangetto, Marco and Sugimoto, Akihiro and Tartaglione, Enzo},
  eprint = {2501.13558v1},
  journal = {arXiv preprint arXiv:2501.13558v1},
  title = {{GoDe: Gaussians on Demand for Progressive Level of Detail and Scalable Compression}},
  url = {https://arxiv.org/abs/2501.13558v1},
  year = {2025}
}

@inproceedings{fan2024lightgaussian,
  author = {Fan, Zhiwen and Wang, Kevin and Wen, Kairun and Zhu, Zehao and Xu, Dejia and Wang, Zhangyang},
  booktitle = {Advances in Neural Information Processing Systems},
  doi = {10.52202/079017-4447},
  pages = {140138--140158},
  title = {{LightGaussian: Unbounded 3D Gaussian Compression with 15x Reduction and 200+ FPS}},
  url = {https://proceedings.neurips.cc/paper_files/paper/2024/file/fd881d3b625437354d4421818f81058f-Paper-Conference.pdf},
  volume = {37},
  year = {2024}
}

@inproceedings{Girish_EAGLES_ECCV2024,
  author = {Girish, Sharath and Gupta, Kamal and Shrivastava, Abhinav},
  booktitle = {Computer Vision -- ECCV 2024},
  doi = {10.1007/978-3-031-73036-8_4},
  pages = {54--71},
  title = {{EAGLES: Efficient Accelerated 3D Gaussians with Lightweight EncodingS}},
  url = {https://doi.org/10.1007/978-3-031-73036-8_4},
  year = {2025}
}

@inproceedings{guo2026matryoshka,
  author = {Guo, Zhilin and Zhang, Boqiao and Aktas, Hakan and Fogarty, Kyle and Hu, Jeffrey and Aslan, Nursena Koprucu and Li, Wenzhao and Baykal, Canberk and Miao, Albert and Bengtson, Josef and Zhou, Chenliang and Xia, Weihao and Vasconcelos, Cristina Nader and Oztireli, Cengiz},
  booktitle = {Computer Vision -- ECCV 2026},
  doi = {10.1007/978-3-032-36984-0_18},
  pages = {323--340},
  title = {{Matryoshka Gaussian Splatting}},
  url = {https://doi.org/10.1007/978-3-032-36984-0_18},
  year = {2026}
}

@inproceedings{hanson2025speedy,
  author = {Hanson, Alex and Tu, Allen and Lin, Geng and Singla, Vasu and Zwicker, Matthias and Goldstein, Tom},
  booktitle = {Proceedings of the IEEE/CVF Conference on Computer Vision and Pattern Recognition (CVPR)},
  month = {June},
  pages = {21537--21546},
  title = {{Speedy-Splat: Fast 3D Gaussian Splatting with Sparse Pixels and Sparse Primitives}},
  year = {2025}
}

@inproceedings{hanson2025pup,
  address = {Nashville, TN, USA},
  author = {Hanson, Alex and Tu, Allen and Singla, Vasu and Jayawardhana, Mayuka and Zwicker, Matthias and Goldstein, Tom},
  booktitle = {Proceedings of the IEEE/CVF Conference on Computer Vision and Pattern Recognition (CVPR)},
  pages = {5949--5958},
  publisher = {IEEE},
  title = {{PUP 3D-GS: Principled Uncertainty Pruning for 3D Gaussian Splatting}},
  year = {2025}
}

@article{DeepBlending2018,
  author = {Hedman, Peter and Philip, Julien and Price, True and Frahm, Jan-Michael and Drettakis, George and Brostow, Gabriel},
  doi = {10.1145/3272127.3275084},
  journal = {ACM Transactions on Graphics},
  number = {6},
  pages = {257:1--257:15},
  publisher = {ACM},
  title = {{Deep Blending for Free-Viewpoint Image-Based Rendering}},
  url = {https://doi.org/10.1145/3272127.3275084},
  volume = {37},
  year = {2018}
}

@article{kerbl3Dgaussians,
  author = {Kerbl, Bernhard and Kopanas, Georgios and Leimk{\"u}hler, Thomas and Drettakis, George},
  doi = {10.1145/3592433},
  journal = {ACM Transactions on Graphics},
  month = {July},
  number = {4},
  pages = {139:1--139:14},
  title = {{3D Gaussian Splatting for Real-Time Radiance Field Rendering}},
  url = {https://doi.org/10.1145/3592433},
  volume = {42},
  year = {2023}
}

@article{Knapitsch2017,
  articleno = {78},
  author = {Arno Knapitsch and Jaesik Park and Qian-Yi Zhou and Vladlen Koltun},
  doi = {10.1145/3072959.3073599},
  journal = {ACM Transactions on Graphics},
  number = {4},
  pages = {78:1--78:13},
  title = {{Tanks and Temples: Benchmarking Large-Scale Scene Reconstruction}},
  url = {https://doi.org/10.1145/3072959.3073599},
  volume = {36},
  year = {2017}
}

@inproceedings{kulhanek2025lodge,
  author = {Kulhanek, Jonas and Rakotosaona, Marie-Julie and Manhardt, Fabian and Tsalicoglou, Christina and Niemeyer, Michael and Sattler, Torsten and Peng, Songyou and Tombari, Federico},
  booktitle = {Advances in Neural Information Processing Systems},
  doi = {10.52202/085713-5474},
  pages = {164054--164075},
  title = {{LODGE: Level-of-Detail Large-Scale Gaussian Splatting with Efficient Rendering}},
  url = {https://proceedings.neurips.cc/paper_files/paper/2025/file/f00475398609746d67821a02765eb8e8-Paper-Conference.pdf},
  volume = {38, Main Conference},
  year = {2025}
}

@inproceedings{lee2024c3dgs,
  address = {Seattle, WA, USA},
  author = {Lee, Joo Chan and Rho, Daniel and Sun, Xiangyu and Ko, Jong Hwan and Park, Eunbyung},
  booktitle = {Proceedings of the IEEE/CVF Conference on Computer Vision and Pattern Recognition (CVPR)},
  pages = {21719--21728},
  publisher = {IEEE},
  title = {{Compact 3D Gaussian Representation for Radiance Field}},
  year = {2024}
}

@inproceedings{Liu_2025_CVPR,
  author = {Liu, Hengyu and Wang, Yuehao and Li, Chenxin and Cai, Ruisi and Wang, Kevin and Li, Wuyang and Molchanov, Pavlo and Wang, Peihao and Wang, Zhangyang},
  booktitle = {Proceedings of the IEEE/CVF Conference on Computer Vision and Pattern Recognition (CVPR)},
  month = {June},
  pages = {16336--16345},
  title = {{FlexGS: Train Once, Deploy Everywhere with Many-in-One Flexible 3D Gaussian Splatting}},
  year = {2025}
}

@inproceedings{liu2025maskgaussian,
  address = {Nashville, TN, USA},
  author = {Liu, Yifei and Zhong, Zhihang and Zhan, Yifan and Xu, Sheng and Sun, Xiao},
  booktitle = {Proceedings of the IEEE/CVF Conference on Computer Vision and Pattern Recognition (CVPR)},
  pages = {681--690},
  publisher = {IEEE},
  title = {{MaskGaussian: Adaptive 3D Gaussian Representation from Probabilistic Masks}},
  year = {2025}
}

@inproceedings{scaffoldgs,
  author = {Lu, Tao and Yu, Mulin and Xu, Linning and Xiangli, Yuanbo and Wang, Limin and Lin, Dahua and Dai, Bo},
  booktitle = {Proceedings of the IEEE/CVF Conference on Computer Vision and Pattern Recognition (CVPR)},
  pages = {20654--20664},
  title = {{Scaffold-GS: Structured 3D Gaussians for View-Adaptive Rendering}},
  year = {2024}
}

@inproceedings{niedermayr2024compressed,
  author = {Niedermayr, Simon and Stumpfegger, Josef and Westermann, R\"udiger},
  booktitle = {Proceedings of the IEEE/CVF Conference on Computer Vision and Pattern Recognition (CVPR)},
  month = {June},
  pages = {10349--10358},
  title = {{Compressed 3D Gaussian Splatting for Accelerated Novel View Synthesis}},
  year = {2024}
}

@inproceedings{Pan2026learning,
  author = {Pan, Youqi and Zhou, Wugen and Zha, Hongbin},
  booktitle = {Proceedings of the IEEE/CVF Conference on Computer Vision and Pattern Recognition (CVPR)},
  month = {June},
  pages = {40939--40948},
  title = {{Learning Differentiable Hierarchies in 3D Gaussian Splatting}},
  year = {2026}
}

@article{octree,
  address = {Los Alamitos, CA, USA},
  author = {Ren, Kerui and Jiang, Lihan and Lu, Tao and Yu, Mulin and Xu, Linning and Ni, Zhangkai and Dai, Bo},
  doi = {10.1109/tpami.2025.3568201},
  issn = {1939-3539},
  journal = {IEEE Transactions on Pattern Analysis and Machine Intelligence},
  month = {May},
  pages = {1--15},
  publisher = {IEEE Computer Society},
  title = {{Octree-GS: Towards Consistent Real-time Rendering with LOD-Structured 3D Gaussians}},
  url = {https://doi.org/10.1109/tpami.2025.3568201},
  year = {2025}
}

@article{seo2025flod,
  address = {New York, NY, USA},
  articleno = {112},
  author = {Seo, Yunji and Choi, Young Sun and Son, HyunSeung and Uh, Youngjung},
  doi = {10.1145/3731430},
  issn = {0730-0301},
  issue_date = {August 2025},
  journal = {ACM Transactions on Graphics},
  month = {July},
  number = {4},
  numpages = {16},
  pages = {112:1--112:16},
  publisher = {Association for Computing Machinery},
  title = {{FLoD: Integrating Flexible Level of Detail into 3D Gaussian Splatting for Customizable Rendering}},
  url = {https://doi.org/10.1145/3731430},
  volume = {44},
  year = {2025}
}

@inproceedings{Shen_2025_CVPR,
  author = {Shen, Jianxiong and Qian, Yue and Zhan, Xiaohang},
  booktitle = {Proceedings of the IEEE/CVF Conference on Computer Vision and Pattern Recognition (CVPR)},
  month = {June},
  pages = {671--680},
  title = {{LOD-GS: Achieving Levels of Detail using Scalable Gaussian Soup}},
  year = {2025}
}

@inproceedings{shi2025lapisgs,
  author = {Shi, Yuang and Morin, G{\'e}raldine and Gasparini, Simone and Ooi, Wei Tsang},
  booktitle = {2025 International Conference on 3D Vision (3DV)},
  doi = {10.1109/3dv66043.2025.00096},
  organization = {IEEE},
  pages = {991--1000},
  title = {{LapisGS: Layered Progressive 3D Gaussian Splatting for Adaptive Streaming}},
  url = {https://doi.org/10.1109/3dv66043.2025.00096},
  year = {2025}
}

@inproceedings{tsai2025l3gs,
  author = {Tsai, Yi-Zhen and Zhang, Xuechen and Li, Zheng and Chen, Jiasi},
  booktitle = {Proceedings of the 31st Annual International Conference on Mobile Computing and Networking},
  doi = {10.1145/3680207.3723485},
  pages = {453--467},
  title = {{L3GS: Layered 3D Gaussian Splats for Efficient 3D Scene Delivery}},
  url = {https://doi.org/10.1145/3680207.3723485},
  year = {2025}
}

@inproceedings{wang2024rdogaussian,
  author = {Wang, Henan and Zhu, Hanxin and He, Tianyu and Feng, Runsen and Deng, Jiajun and Bian, Jiang and Chen, Zhibo},
  booktitle = {Computer Vision -- ECCV 2024},
  doi = {10.1007/978-3-031-73636-0_5},
  pages = {76--92},
  title = {{End-to-End Rate-Distortion Optimized 3D Gaussian Representation}},
  url = {https://doi.org/10.1007/978-3-031-73636-0_5},
  year = {2025}
}

@inproceedings{Wang_2025_CVPR,
  author = {Wang, Peihao and Wang, Yuehao and Wang, Dilin and Mohan, Sreyas and Fan, Zhiwen and Wu, Lemeng and Cai, Ruisi and Yeh, Yu-Ying and Wang, Zhangyang and Liu, Qiang and Ranjan, Rakesh},
  booktitle = {Proceedings of the IEEE/CVF Conference on Computer Vision and Pattern Recognition (CVPR)},
  month = {June},
  pages = {26663--26672},
  title = {{Steepest Descent Density Control for Compact 3D Gaussian Splatting}},
  year = {2025}
}

@inproceedings{Wang_ContextGS_NeurIPS2024,
  author = {Wang, Yufei and Li, Zhihao and Guo, Lanqing and Yang, Wenhan and Kot, Alex C. and Wen, Bihan},
  booktitle = {Advances in Neural Information Processing Systems},
  doi = {10.52202/079017-1631},
  pages = {51532--51551},
  title = {{ContextGS : Compact 3D Gaussian Splatting with Anchor Level Context Model}},
  url = {https://proceedings.neurips.cc/paper_files/paper/2024/file/5c20ca4b0b20b0bd2f1d839dc605e70f-Paper-Conference.pdf},
  volume = {37},
  year = {2024}
}

@article{wang2004ssim,
  author = {Wang, Zhou and Bovik, Alan C. and Sheikh, Hamid R. and Simoncelli, Eero P.},
  doi = {10.1109/tip.2003.819861},
  journal = {IEEE Transactions on Image Processing},
  number = {4},
  pages = {600--612},
  title = {{Image Quality Assessment: From Error Visibility to Structural Similarity}},
  url = {https://doi.org/10.1109/tip.2003.819861},
  volume = {13},
  year = {2004}
}

@inproceedings{yang2026gs2,
  author = {Yang, Xianben and Wang, Tao and Li, Yuxuan and Jin, Yi and Ling, Haibin},
  booktitle = {Proceedings of the IEEE/CVF Conference on Computer Vision and Pattern Recognition (CVPR)},
  month = {June},
  pages = {33259--33268},
  title = {{{GS$^2$}: Graph-based Spatial Distribution Optimization for Compact 3D Gaussian Splatting}},
  year = {2026}
}

@inproceedings{zhang2018lpips,
  author = {Zhang, Richard and Isola, Phillip and Efros, Alexei A. and Shechtman, Eli and Wang, Oliver},
  booktitle = {Proceedings of the IEEE Conference on Computer Vision and Pattern Recognition (CVPR)},
  doi = {10.1109/CVPR.2018.00068},
  pages = {586--595},
  title = {{The Unreasonable Effectiveness of Deep Features as a Perceptual Metric}},
  year = {2018}
}

@inproceedings{zhang2025gaussianspa,
  author = {Zhang, Yangming and Jia, Wenqi and Niu, Wei and Yin, Miao},
  booktitle = {Proceedings of the IEEE/CVF Conference on Computer Vision and Pattern Recognition (CVPR)},
  month = {June},
  pages = {26673--26682},
  title = {{{GaussianSpa}: An ``Optimizing-Sparsifying'' Simplification Framework for Compact and High-Quality 3D Gaussian Splatting}},
  year = {2025}
}

@inproceedings{zhang2024lp,
  author = {Zhang, Zhaoliang and Song, Tianchen and Lee, Yongjae and Yang, Li and Peng, Cheng and Chellappa, Rama and Fan, Deliang},
  booktitle = {Advances in Neural Information Processing Systems},
  doi = {10.52202/079017-3891},
  pages = {122434--122457},
  title = {{LP-3DGS: Learning to Prune 3D Gaussian Splatting}},
  url = {https://proceedings.neurips.cc/paper_files/paper/2024/file/dd51dbce305433cd60910dc5b0147be4-Paper-Conference.pdf},
  volume = {37},
  year = {2024}
}

@inproceedings{Zoomers_2025_WACV,
  author = {Zoomers, Brent and Wijnants, Maarten and Molenaers, Ivan and Vanherck, Joni and Put, Jeroen and Jorissen, Lode and Michiels, Nick},
  booktitle = {Proceedings of the Winter Conference on Applications of Computer Vision (WACV)},
  month = {February},
  pages = {3118--3127},
  title = {{PRoGS: Progressive Rendering of Gaussian Splats}},
  year = {2025}
}
